\documentclass[10pt,twocolumn,letterpaper]{article}

\usepackage[pagenumbers]{iccv} 

\newcommand{\blue}[1]{{\color{blue}#1}}

\definecolor{iccvblue}{rgb}{0.21,0.49,0.74}
\usepackage[pagebackref,breaklinks,colorlinks,allcolors=iccvblue]{hyperref}
\usepackage{multirow}
\usepackage{multicol}
\usepackage{pifont}
\usepackage{amsmath}
\usepackage{epigraph}
\usepackage{tcolorbox}
\usepackage{etoolbox}
\usepackage{float}
\usepackage{xr}
\definecolor{iccvblue}{rgb}{0.21,0.49,0.74}
\usepackage[pagebackref,breaklinks,colorlinks,allcolors=iccvblue]{hyperref}
\hypersetup{
    colorlinks=true,
    urlcolor=magenta
}

\newcommand{\ours}{ObjectMLLM\xspace}

\title{How Can Objects Help Video-Language Understanding?}

\author{Zitian Tang$^1$\quad Shijie Wang$^1$\quad Junho Cho$^2$\quad Jaewook Yoo$^2$\quad Chen Sun$^1$\\
$^1$Brown University\quad$^2$Samsung Electronics
\vspace{1mm}\\
\url{https://brown-palm.github.io/ObjectMLLM}
}

\makeatletter
\g@addto@macro\@maketitle{
\vspace{-3em}
\begin{figure}[H]
\setlength{\linewidth}{\textwidth}
\setlength{\hsize}{\textwidth}
\centering
\includegraphics[trim={0cm, 0cm, 0cm, 0.0cm},clip,width=0.85\linewidth]{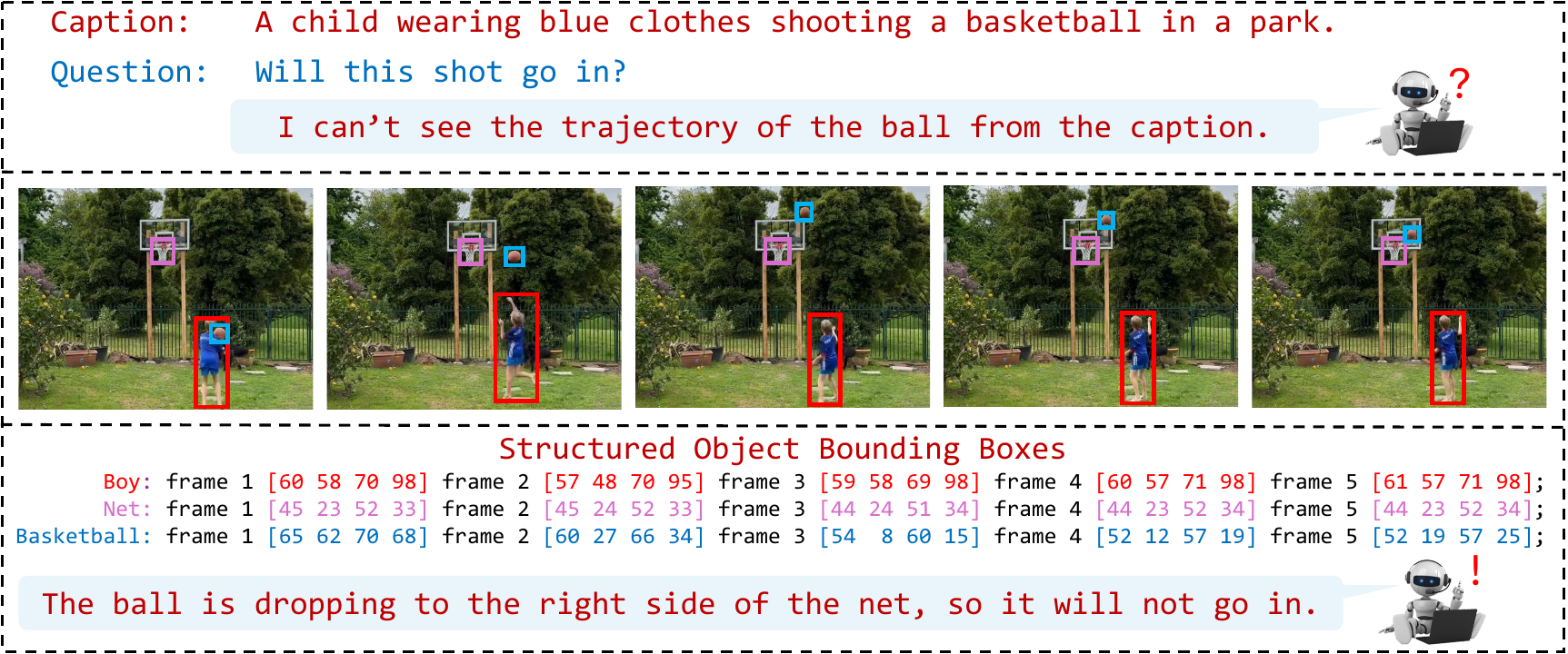}
\vspace{-2.5mm}
\caption{
{Socratic Models~\cite{zeng2022socratic,wang2023vamos,zhang2023simple} perceive the world from the lens of natural language descriptions, which may miss important spatiotemporal information  (\textit{top}). Multimodal large language models (MLLMs), on the other hand, can integrate visual information via distributed embeddings, but typically require large-scale instruction tuning datasets for \textit{adaptation}. We investigate how explicit, continuous object representations (\eg box coordinates from object detectors) can help video-language understanding (\textit{bottom}).
} 
}
\label{fig:teaser}
\end{figure}
}
\makeatother

\begin{document}
\maketitle
\begin{abstract}
Do we still need to represent objects explicitly in multimodal large language models (MLLMs)? To one extreme, pre-trained encoders convert images into visual tokens, with which objects and spatiotemporal relationships may be implicitly modeled. To the other extreme, image captions by themselves provide strong empirical performances for understanding tasks, despite missing fine-grained spatiotemporal information. To answer this question, we introduce ObjectMLLM, a framework capable of leveraging arbitrary computer vision algorithm to extract and integrate structured visual representation. Through extensive evaluations on six video question answering benchmarks, we confirm that explicit integration of object-centric representation remains necessary. Surprisingly, we observe that the simple approach of quantizing the continuous, structured object information and representing them as plain text performs the best, offering a data-efficient approach to integrate other visual perception modules into MLLM design. Our code and models are released at \url{https://github.com/brown-palm/ObjectMLLM}.
\end{abstract}    
\section{Introduction}
\label{sec:intro}

What makes a good representation for video-language understanding? In the era of multimodal large language models (MLLMs), anything that can be \textit{tokenized} has the potential to serve as a valid representation. Along the spectrum are two extremes: those that project arbitrary distributed representations to the input space of a pre-trained large language model via instruction tuning~\cite{liu2023llava,han2023imagebindllmm}, and those that model the visual world as interpretable concepts~\cite{wang2022language} and captions~\cite{wang2023vamos,zhang2023simple}, which can be directly consumed by LLMs via Socratic Methods~\cite{zeng2022socratic}. It is open to debate whether either approach can effectively capture and convey the complexity of the visual world to an LLM ``reasoner''. As illustrated in Figure~\ref{fig:teaser}, video captions tend to ignore detailed information that captures the spatiotemporal object configurations. Meanwhile, despite inductive biases to guide MLLM encoders to be spatial aware~\cite{tong2024cambrian}, integrating visual information such as objects and their locations into LLMs remains a challenging endeavor~\cite{tong2024eyes}.

We hypothesize that \textit{explicit} object-centric recognition and modeling, supported by the rich literature from the computer vision community, remains essential to the success of MLLMs. We then seek to answer the question, \textit{how can objects help video-language understanding in MLLMs}, from two perspectives: representation and adaptation. Motivated by the effectiveness of caption-based representation for video understanding~\cite{wang2023vamos,zhang2023simple,min2024morevqa,wang2024lifelongmemory}, we hypothesize that there is a natural trade-off between the expressiveness of a visual representation, and the easiness for it to be adapted into a pre-trained LLM: A  distributed representation is the most expressive, but needs larger amount of instruction tuning data for it to be integrated into LLMs~\cite{li2023blip,liu2023llava}. ``Symbolic'' representations that are language-based (\eg, rendering quantized object coordinates as plain text), though less expressive, may be easier to use as they can be represented with the existing vocabulary of an LLM. We further hypothesize that symbolic object representations are more expressive in videos when they depict the trajectories of a moving object or its key points, as 
Johansson's biological motion perception experiment~\cite{johansson1973visual} showed that humans can successfully associate a collection of dots with human motions as soon as the dots start moving (\eg, the trajectory of the basketball in Figure~\ref{fig:teaser}).

To validate these hypotheses, we introduce \textbf{ObjectMLLM}, a framework capable of leveraging arbitrary computer vision algorithm (\eg, an object detector or human pose estimator) to extract and integrate structured visual representation into multimodal LLMs. With ObjectMLLM, we investigate the trade-off of designing object-centric representations, either by learning an embedding projector, or with the symbolic object representation. The former approach generates a distributed representation projected into the input space of an LLM, from a vectorized representation of object bounding boxes. The latter approach directly renders bounding boxes as \textit{strings}, which are then tokenized accordingly. For both approaches, we apply parameter efficient fine-tuning to adapt the weights of the pre-trained LLMs  together towards the target tasks. We observe that as hypothesized, while embedding projector leads to more compact object representations, it is less data-efficient compared to symbolic object representation, consistently yielding lower performance when fine-tuned for the same number of iterations. We then conduct thorough evaluations on six video QA benchmarks, where we observe that symbolic object representation consistently improves the model performance, especially on tasks that require spatiotemporal understanding (\eg, PerceptionTest~\cite{patraucean2023perception}).

In summary, our contributions are three-fold:
\begin{itemize}
    \item We propose \ours, a multimodal video understanding framework that seamlessly incorporates object spatial information from computer vision algorithms in multimodal LLMs.
    \item We study two bounding box adapters and show that a language-based representation is more performant and data-efficient than latent embedding projectors, indicating pre-trained LLMs may already be \textit{spatially aware}.
    \item Our evaluation on video question answering benchmarks demonstrates the significance of ObjectMLLM when applied to both pre-trained LLMs and multimodal LLMs, and that the benefits generalize to other structured visual representation, such as human joint coordinates~\cite{endo2023humanmotionqa}.
\end{itemize}

\section{Related Works}

\begin{figure*}[t]
    \centering
    \includegraphics[width=\linewidth]{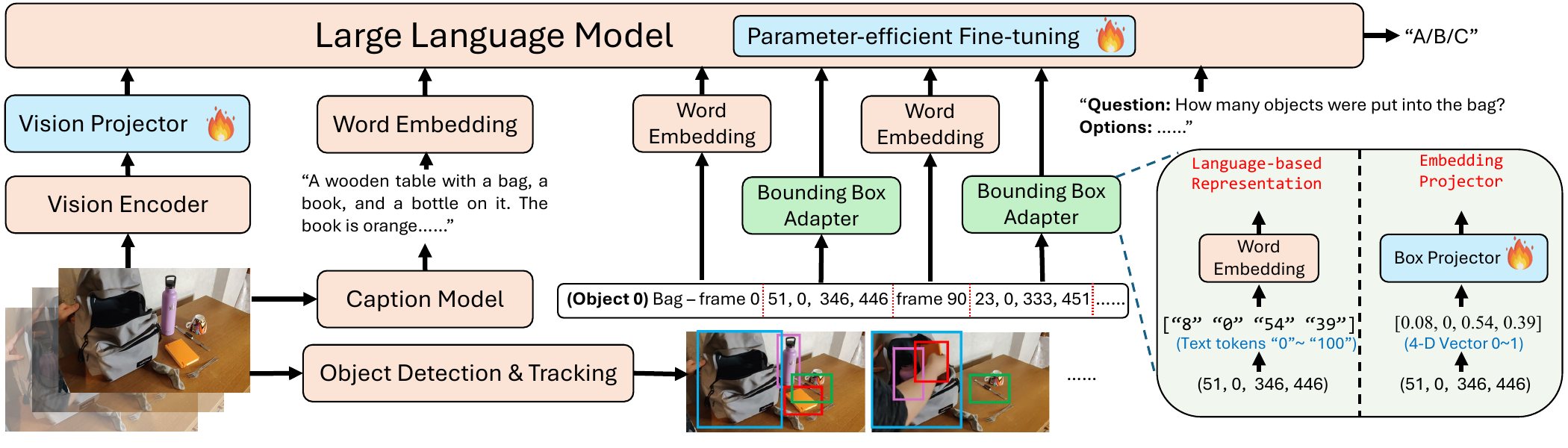}
    \vspace{-15pt}
    \caption{Pipeline of \ours. The possible model inputs are visual embeddings, video frame captions, and object bounding boxes. ObjectMLLM encodes object coordinates with a language-based (``symbolic'') representation, or with an embedding projector. The former represents the bounding boxes as plain text, while the latter maps the vectorized coordinates into the input space of the LLM.}
    \label{fig:framework}
    \vspace{-5pt}
\end{figure*}

\subsection{Video Large Language Models}
Large language models (LLMs) have recently shown remarkable progress in understanding and generating text across various domains. Their success has inspired the development of Video Large Language Models (Video-LLMs)~\cite{jin2023chatunivi, zhang2023video, ko2023large, damonlpsg2024videollama2, xue2024xgen, wang2023vamos, zhao2024antgpt, 2023videochat, yu2023self, li2024llava}, which integrate videos into the language modeling framework and are widely applied in tasks such as video captioning and question answering. Most Video-LLMs consist of three components: a pre-trained visual encoder, an adaptation model, and an LLM backbone. One of the primary challenges for Video-LLMs, compared to Image-LLMs, is how to effectively and efficiently representing the rich contextual information in videos. Many Video-LLMs~\cite{zhang2023video, damonlpsg2024videollama2, ko2023large, xue2024xgen, li2024llava} employ pre-trained image encoders~\cite{Radford2021LearningTV, zhai2023sigmoid, oquab2024dinov} to extract features from sampled frames individually, concatenating them to form video representations. Other approaches~\cite{wang2022internvideo, Maaz2023VideoChatGPT, lin-etal-2024-video} utilize a dedicated video encoder to capture spatial-temporal features across the entire video. Additionally, Chat-UniVi~\cite{jin2023chatunivi} combines image and video encoders and implements spatial merging to reduce the number of video tokens for greater efficiency. Beyond video features, some models, such as Vamos~\cite{wang2023vamos}, VideoChat~\cite{2023videochat}, and LifelongMemory~\cite{wang2024lifelongmemory}, flexibly incorporate action labels and video captions as inputs to represent videos from multiple perspectives. In this work, we investigate the influence of object-centric information in Video-LLMs and explore methods to incorporate structured representations, such as objects represented by sequences of bounding boxes and class labels, into Video-LLMs.

\subsection{Modality adaptation in MLLMs}
Modality adaptation in multimodal large language models (MLLMs) is critical when extending large language models to handle diverse inputs, including images, audio, and video.  One intuitive approach for non-text modalities is to convert them into textual representations, such as captions~\cite{zeng2022socratic, wang2023vamos, zhang2023simple, berrios2023towards} or action labels~\cite{zhao2024antgpt}. Such textual representations provide good interpretability and data efficiency by leveraging the extensive language prior knowledge embedded in LLMs. Through this method, domain-specific expert models, such as video captioning and action recognition models, act as adaptation modules within the multimodal LLM framework. Another common approach for aligning non-text modalities to the text space is multimodal fine-tuning, which directly uses continuous embeddings and trains a projection module for adaptation. Two types of projection modules are frequently employed: MLP projectors and attention-based projectors~\cite{li2023blip}. For instance, LLaVA~\cite{liu2023llava} utilizes a lightweight linear layer to project vision embeddings to input token for the LLM through multi-stage training on large-scale datasets, while LLaVA1.5~\cite{Liu_2024_CVPR} further improves by adopting a two-layer MLP projector. Recent studies~\cite{lin2024vila, mckinzie2024mm1} suggest that the specific structure of the projector exerts marginal influence on MLLM performance. Compared with textual representations, multimodal fine-tuning directly utilize continuous embeddings from encoders but generally requires substantial multi-stage training on large-scale multimodal datasets. In this work, we systematically compare various approaches for adapting structured object representations within Video-LLMs and evaluate the impact of different modality representations on video question answering tasks.

\subsection{Objects in MLLMs}

A prominent approach to integrate objects into MLLMs involves leveraging object detectors to extract region-based features for downstream tasks. OSCAR~\cite{li2020oscar} introduces an object-aware pre-training paradigm that aligns object tags with textual data, enhancing contextual understanding. VinVL~\cite{zhang2021vinvl} builds upon OSCAR by employing a stronger object detector to extract more accurate region features. CoVLM~\cite{li2023covlm} advances this direction by explicitly composing visual entities and relationships within text through the use of communication tokens. These tokens facilitate dynamic interaction between the visual detection system and the language system. When communication tokens are generated by the LLM, detection models respond by generating regions-of-interest (ROIs), which are then fed back into the LLM to improve language generation. Another line of work focuses on grounding VLMs, which are capable of localizing objects and predict bounding boxes or masks based on language references. Models such as Shikra~\cite{chen2023shikra}, Kosmos-2~\cite{peng2023kosmos}, and GLaMM~\cite{rasheed2024glamm} are trained on large scale grounding and localization dataset, and are designed specifically for these tasks. In these models, structured localization information such as bounding boxes is usually encoded and projected to align with LLMs and a decoding head is trained to make prediction. ObjectMLLM aligns with the first approach by investigating how object-centric information can enhance video understanding in multimodal LLMs.
\section{Method}

We aim to complement Multimodal Large Language Models (MLLMs) with structured visual information extracted by off-the-shelf computer vision algorithms. We focus on object-centric representation, which captures object location and motion.
As \Cref{fig:teaser} shows, the object-centric representation encodes the position and movements of individual objects via 2D bounding boxes, object labels, and timestamps.
We expect that enabling MLLMs to explicitly utilize object-centric representation can enhance their spatiotemporal understanding capability.
For this purpose, we investigate whether and how we can boost video understanding by leveraging object bounding boxes.

Our investigation focuses on two perspectives: The first is the final video question answering accuracy, measuring how useful an object-centric representation is; the second is the amount of fine-tuning data required for a pre-trained LLM or multimodal LLM to utilize the object information properly, which we refer to as  \textit{data efficiency}.
Both have practical motivations: as it is desirable to explore the explicit integration of different object detectors and trackers, or even computer vision models for pose estimation, panoptic segmentation, into LLMs -- without the need to always perform large-scale instruction tuning. We are also interested in a more philosophical discussion on to what degree LLMs pre-trained on language data are \textit{spatially aware}, and whether they can be tuned to perform spatial understanding in a data-efficient manner.

\subsection{\ours}

We propose \ours, a multimodal framework that integrates distributed visual embeddings, video frame captions, and object bounding boxes into one MLLM, as illustrated in \Cref{fig:framework}.
The utilization of video frame embeddings and captions is in line with caption-enhanced MLLMs, \eg, Vamos \cite{wang2023vamos}.
Specifically, we uniformly sample a fixed number of frames from a video, and employ an off-the-self image feature encoder and captioning model to extract visual embeddings and captions, respectively.
The generated captions are directly fed to the LLM as inputs, while the visual embeddings are mapped into the word embedding space of the LLM by a vision projector, typically implemented as a lightweight neural network.

\begin{figure}[t]
    \centering
    \includegraphics[width=\linewidth]{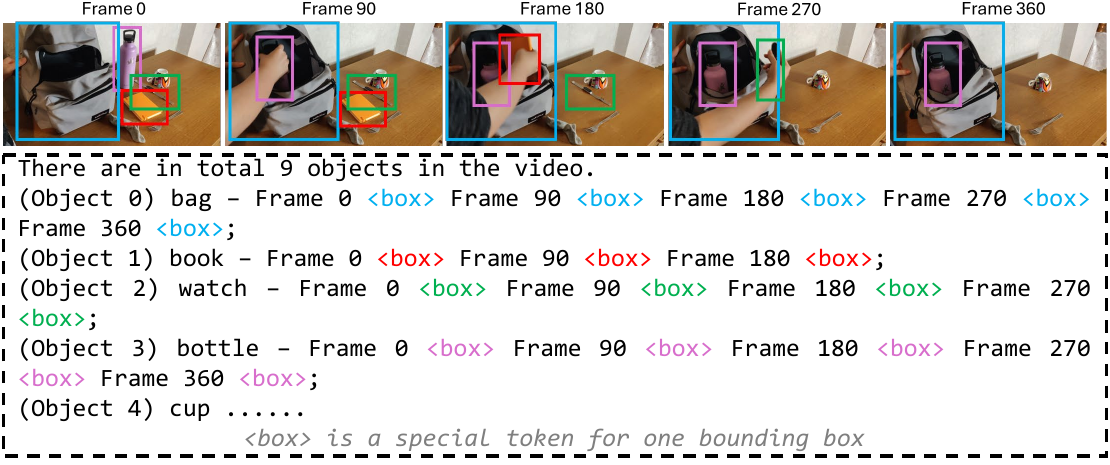}
    \vspace{-15pt}
    \caption{Template to format the object bounding boxes. The timestamps when an object is visible are listed, each of which is followed by tokens that describe the bounding box coordinates.}
    \label{fig:bbox_template}
    \vspace{-15pt}
\end{figure}

With off-the-shelf object detection and tracking models, we capture object bounding boxes from the video.
Following the template in \Cref{fig:bbox_template}, we list the timestamps when each object is visible, and append a special bounding box token after each timestamp.
The textual part, including the object labels and timestamps, are directly tokenized and converted to word embeddings by the LLM. Each bounding box, which is represented by four continuous numbers, can either be rendered as plain text and tokenized as text tokens, or passed to a projector to produce an embedding in the LLM input space. The bounding box tokens are interleaved with the text tokens corresponding to the object labels and timestamps. In Section~\ref{sec:method_adapter}, we discuss the strengths and limitations of each bounding box representation.

\begin{figure}[t]
    \centering
    \includegraphics[width=\linewidth]{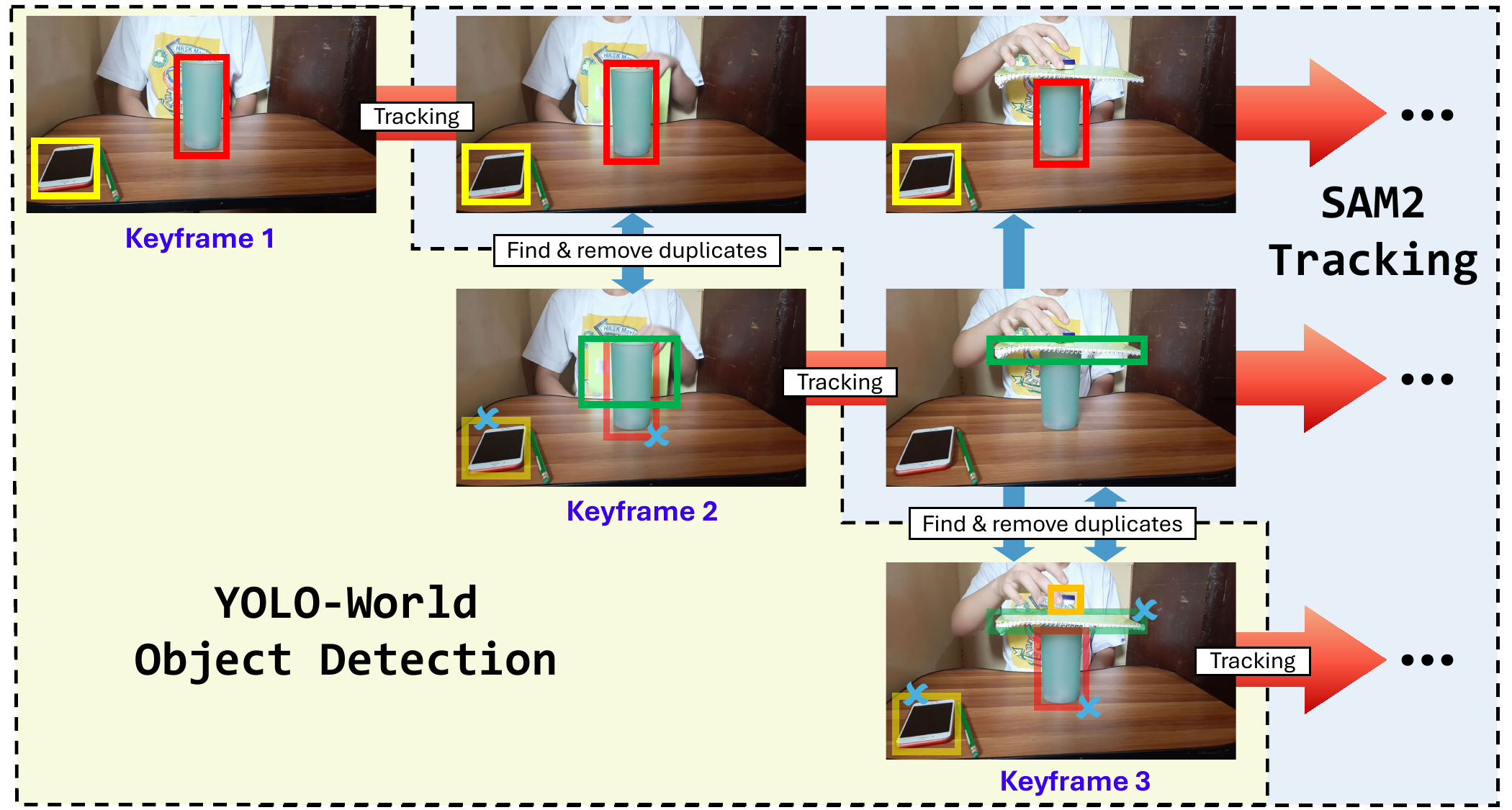}
    \caption{Workflow of object detection and tracking.  We iteratively detect objects in uniformly sampled keyframes, and create new tracks for new objects not associated with any existing track.}
    \label{fig:tracking_workflow}
    \vspace{-15pt}
\end{figure}

\subsection{Object detection and tracking}
To obtain object-centric representations, we need the semantic labels and tracked bounding boxes of the objects in a video.
The computer vision community has developed powerful, standalone models for object detection and tracking. We choose YOLO-World~\cite{Cheng2024YOLOWorld}, an open-vocabulary object detector, to detect objects in a given video frame, and use SAM 2 \cite{ravi2024sam2} to track the detected objects across the video.

The workflow is illustrated in \Cref{fig:tracking_workflow}. For each video benchmark, we consider all the object categories in its training set as the vocabulary of YOLO-World, which detects objects from video frames that are uniformly sampled. For each subsequent frame after the initial one, we deploy both SAM 2 to track objects already detected in the previous frames, and also YOLO-World to detect all objects present in the current frame. We calculate the IoU between the detected objects and the objects from existing tracks. Detected objects with an IoU greater than $0.5$ are removed as duplicates, and the remaining ones are used to create new tracks.

To mitigate the distribution shift compared to its pre-training data, YOLO-World is fine-tuned on the training set of each benchmark, respectively, before usage. The pre-trained
SAM 2 is kept frozen for all benchmarks.

\subsection{Object-centric representation}
\label{sec:method_adapter}

As illustrated in \Cref{fig:framework}, we consider two implementations of bounding box adapters. The language-based adapter turns continuous boxes into interpretable symbols that represent the spatial locations, and  the embedding projector learns to project the vectorized bounding box coordinates of arbitrary object into the LLM input space. Intuitively, the language-based representation could be more data-efficient since it directly reuses the tokenizer and word embeddings from a pre-trained (multimodal) LLM.

\vspace{.3em}\noindent\textbf{Language-based representation:}
We perform normalization and quantization to map continuous bounding box coordinates into discrete integers. This conversion is lossy but uses fewer tokens than float numbers. We take values in the range of $[0,100]$. Each value is tokenized and embedded with the existing LLM tokenizer.
A drawback of this method is that it uses multiple tokens to represent one bounding box, requiring long context windows of the LLM, and is computationally more expensive.

\vspace{.3em}\noindent\textbf{Embedding projector:}
A commonly-adopted approach for a pre-trained MLLM to incorporate non-textual information is to train a embedding projector that maps the vectorized representation from a new data ``modality'' to the input space of the LLM.
For example, LLaVA~\cite{liu2023llava} trains a linear layer as the projector of image CLIP embeddings. \ours takes a bounding box as a 4-dimensional embedding with continuous values, normalizes each dimension to floats in $\left[0,1\right]$, and trains a linear layer as the embedding projector to produce vectorized representations with the same number of the dimensions as the LLM word embeddings.

\subsection{Fine-tuning strategy}
\label{sec:finetune_strategy}

\ours can be fine-tuned from either a pre-trained LLM or a multimodal LLM. 
We perform parameter-efficient fine-tuning on the LLM backbone, and jointly train the vision projector and the embedding projector of the bounding box adapter with the LLM backbone. All other parameters are kept frozen.

When starting from pre-trained LLMs, we adopt a modality-by-modality training strategy used by VideoLLaMA2 \cite{damonlpsg2024videollama2} to gradually incorporate incoming modalities.
For example, to develop a model that incorporates both the caption and the bounding box modality, we first train the model in a caption-only setting.
After the model learns to utilize video frame captions, we further fine-tune it with inputs containing both captions and boxes. The modality incorporation order we use is frame captions, bounding boxes, and visual embeddings across all the benchmarks.
\section{Experiments}

We first compare the two bounding box adapters in \ours.
We then choose the best-performing adapter, and investigate the effectiveness of integrating different input modalities. Finally, we study if object-centric representation can be used to improve the performance of MLLMs that are pre-trained to utilize visual embedding without explicit object-centric representations.

\subsection{Benchmarks}

To evaluate a model's understanding about object bounding boxes, we need benchmarks where spatial and temporal object information is essential to the questions.
CLEVRER \cite{CLEVRER2020ICLR} is a synthetic video dataset focusing on object motion and collision.
However, CLEVRER contains open-ended questions, making the performance measurement difficult.
MVBench \cite{li2023mvbench} converts some of the CLEVRER \cite{CLEVRER2020ICLR} questions into multi-choice questions.
We use this part of data and name it CLEVRER-MC.
To train our models on CLEVRER, we use the CLEVRER-sourced part of VideoChat2-IT \cite{li2023mvbench}. It is also multi-choice questions but may have different question types from CLEVRER-MC.

Besides, we also evaluate the models on real-world video benchmarks --
Perception Test \cite{patraucean2023perception}, STAR \cite{wu2021star_situated_reasoning}, NExT-QA \cite{xiao2021next}, and IntentQA \cite{Li_2023_ICCV}.
While some questions in these benchmarks are related to spatiotemporal object motion, there are also questions focusing on causal reasoning.
Evaluation on these benchmarks can reveal the scenarios where the object-centric representation can make a difference.

    \begin{figure}[t]
        \centering
        \includegraphics[width=1\linewidth]{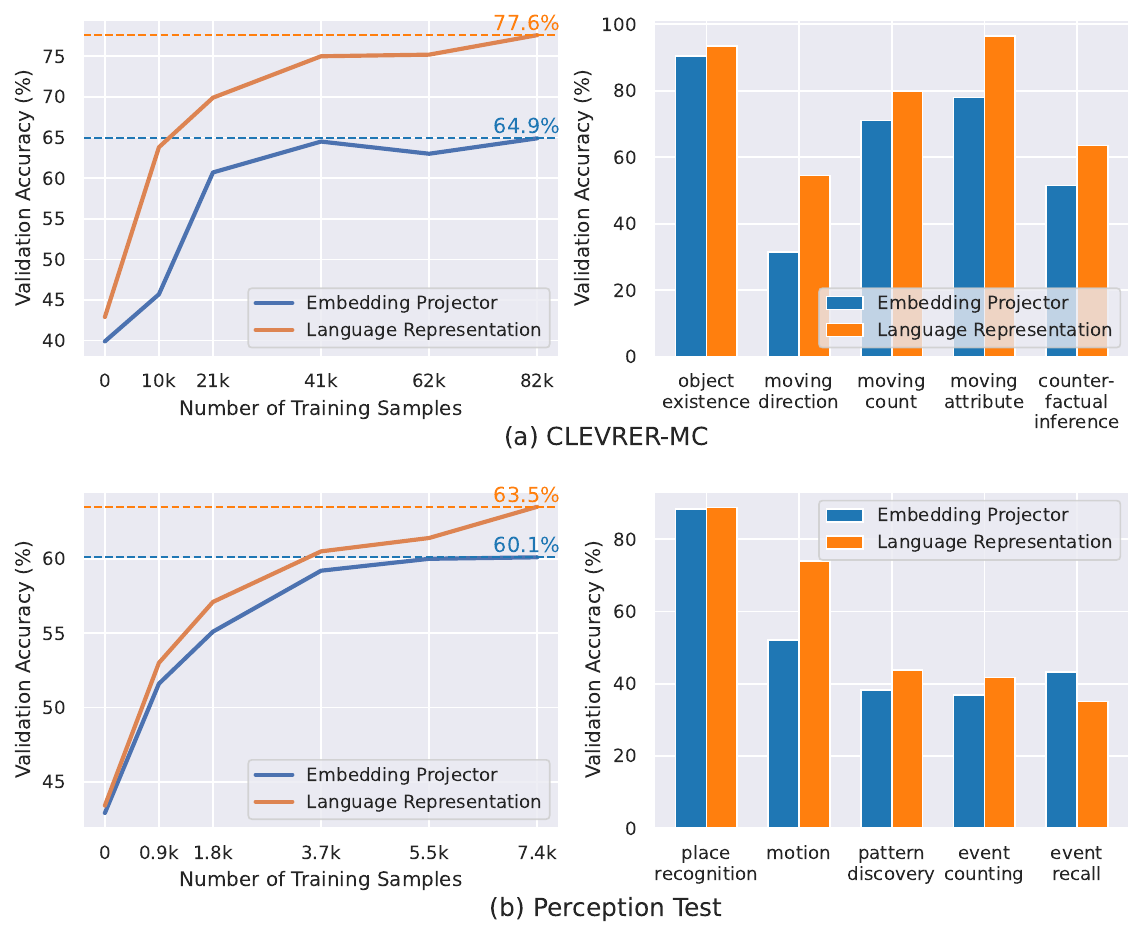}
        \vspace{-20pt}
        \caption{Performance of the box adapters under various training data amounts (left) and accuracy breakdown by question types (right). Only a subset of the question types in Perception Test are listed here. The language-based representation consistently outperforms the embedding projector with different numbers of training samples on both CLEVRER-MC and Perception Test, showing its effectiveness and data efficiency. In the breakdown, the language-based representation outperforms the embedding projector on motion-related questions by large margins. }
        \label{fig:data_efficiency}
        \vspace{-15pt}
    \end{figure}

    \begin{table*}[t]
    \centering
    \scalebox{0.9}{
        \begin{tabular}{ccc|ccccc}
            \toprule
            Video & Caption & Box & CLEVRER-MC & Perception Test & STAR & NExT-QA & IntentQA  \\
            \cmidrule(r){1-8}
            \checkmark & & & 40.3 & 59.6 & 59.7 & 70.7 & 68.2 \\ 
            & \checkmark & & 47.8 & 62.4 & 60.1 & \textbf{76.6} & \textbf{75.7} \\ 
            & & \checkmark & \textbf{77.6} & 63.5 & 59.1 & 63.7 & 66.2 \\ 
            \checkmark & & \checkmark & 77.1 & 62.7 & 59.3 & 71.8 & 71.7 \\ 
            & \checkmark & \checkmark & 75.5 & \textbf{65.7} & \textbf{64.4} & \textbf{76.6} & 75.6 \\ 
            \checkmark & \checkmark & \checkmark &  75.4 & 63.9 & 62.9 & 76.2 & 75.0 \\ 
            \bottomrule
        \end{tabular}}
    \vspace{-5pt}
    \caption{Accuracy under different combinations of modalities. Using object bounding boxes improves the performance on CLEVRER-MC, Perception Test, and STAR by large margins. Caption remains the most effective modality on NExT-QA and IntentQA.}
    \label{tab:ablation_modality}
    \vspace{-10pt}
    \end{table*}

    \begin{table}[t]
    \centering
        \scalebox{0.81}{
        \begin{tabular}{ccc|ccccccc}
            \toprule
            Video & Caption & Box & OE & MD & MC & MA & CI & All \\
            \cmidrule(r){1-9}
            \checkmark & & & 51.0 & 21.0 & 44.5 & 37.0 & 48.0 & 40.3 \\ 
            & \checkmark & & 62.5 & 26.5 & 50.5 & 50.0 & 49.5 & 47.8 \\ 
            & & \checkmark & \textbf{93.5} & \textbf{54.5} & 80.0 & 96.5 & \textbf{63.5} & \textbf{77.6} \\ 
            & \checkmark & \checkmark & 92.5 & 51.0 & 79.0 & \textbf{97.0} & 58.0 & 75.5 \\ 
            \checkmark & \checkmark & \checkmark & 92.0 & 47.5 & \textbf{81.0} & 95.5 & 61.0 & 75.4 \\ 
            \bottomrule
        \end{tabular}
        }
    \vspace{-5pt}
    \caption{Accuracy of different question types on CLEVRER. While the bounding boxes boost the performance across all the question types, it is more significant on OE, MC, and MA than on others. OE: object existence; MD: moving direction; MC: moving count; MA: moving attribute; CI: conterfactual inference.}
    \label{tab:breakdown_clevrer}
    \vspace{-15pt}
    \end{table}
    
\subsection{Implementation}
\label{sec:implementation}

When starting from a pre-trained LLM to build \ours, we follow Vamos \cite{wang2023vamos} and use LLaMA3-8B \cite{grattafiori2024llama3herdmodels} as the backbone and fine-tune it with LLaMA-Adapter \cite{zhang2023llamaadapter}.
The vision projector and box projector in the bounding box adapter are linear layers.
The distributed visual embedding is extracted by CLIP ViT-L/14 \cite{Radford2021LearningTV} on 10 uniformly sampled frames per video.
The embedding projector weights are all initialized to zero, which we find to lead to better performance than the default random initialization in PyTorch.
Moreover, we use LLaVA-1.5-13B \cite{Liu_2024_CVPR} to generate captions for 6 uniformly sampled frames from each video.

When starting from a pre-trained  \emph{multimodal} LLM, we use VideoLLaMA2-7B \cite{damonlpsg2024videollama2}, which is pre-trained on 100M video-language data.
It includes CLIP ViT-L/14 \cite{Radford2021LearningTV} as vision encoder, Spatial-Temporal Convolution as vision projector, and Mistral-7B-Instruct \cite{jiang2023mistral7b} as LLM backbone.
We fine-tune it with LoRA \cite{hu2022lora} in our experiments.

To keep the context length under control, we uniformly sample the object bounding boxes at a lower frame rate such that the language-based representation of all the boxes in a video contains fewer than $1,000$ tokens.
As different videos have different lengths and numbers of objects, the downsampling rate varies from video to video.

The hyperparameters for fine-tuning, bounding box downsampling rates, and implementation details of object detection and tracking are in \Cref{app:detail}.

\subsection{Comparison of adaptation methods}
\label{sec:compare_adapter}

We first compare the two adapters on object-centric representations. For this purpose, we do not use the visual embeddings and captions, and train the model only with the object bounding boxes as input. We focus on the
CLEVRER-MC and Perception Test benchmarks, as we empirically observe that they were designed to contain questions more closely related to spatiotemporal object configurations.

In \Cref{fig:data_efficiency} (left), we evaluate the two adapters with various portions of the training data.
With the full training data, the language-based representation outperforms the embedding projector across both benchmarks ($77.6\%$ \emph{vs.} $64.9\%$ on CLEVRER-MC and $63.5\%$ \emph{vs.} $60.1\%$ on Perception Test).
More importantly, the language-based representation can outperform the embedding projector with \textit{any amount of data}.
Notably, with only one-eighth (10k) of the training data on CLEVRER-MC, the fine-tuned model is able to utilize bounding boxes from language-based representation and achieves an accuracy of $63.8\%$, but the performance with the embedding projector remains low ($44.5\%$).
Although the embedding projector can keep the continuity of the bounding box coordinates, our experiments show that the LLM backbone struggles to understand the resulting box embeddings when limited fine-tuning data is used. Reusing the existing LLM vocabulary, which is done by the language-based representation, lead to effective and data-efficient understanding of the bounding boxes.

\Cref{fig:data_efficiency} (right) shows the accuracy breakdown by question types.
While the performances of the two adapters are comparable on some types of question, the language-based representation shows great superiority on motion-related questions.
This phenomenon in motion questions happens to be consistent with Johansson's biological motion perception experiment~\cite{johansson1973visual} that humans can associate a collection of moving dots with human motions.

    \begin{figure*}[t]
        \centering
        \includegraphics[width=\linewidth]{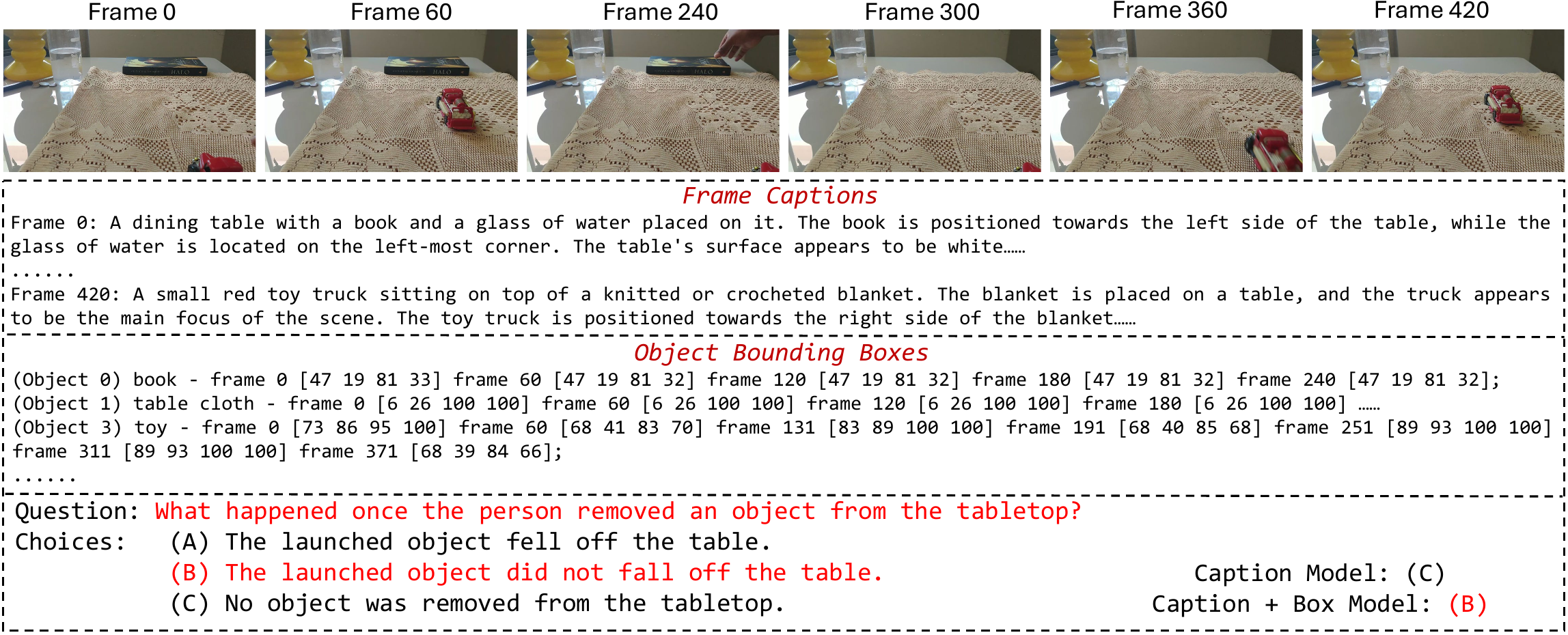}
        \vspace{-20pt}
        \caption{Qualitative example on Perception Test. Although the captions can capture the toy truck on the table, only the caption-and-box model can recognize the spatial relation between the toy truck and the table based on the object bounding boxes.}
        \label{fig:q_ptest}
        \vspace{-5pt}
    \end{figure*}

    \begin{table*}[t]
    \centering
        \scalebox{0.9}{
        \begin{tabular}{c|c|cc|ccccc}
            \toprule
            Setting & Models & Video & Box & CLEVRER-MC & Perception Test & STAR & NExT-QA & IntentQA \\
            \midrule
            \multirow{2}{*}{Zero-shot} & VideoLLaMA2 & \checkmark & & 45.6 & 51.4 & 57.1 & 74.1 & 73.8 \\
            & \ours & \checkmark & \checkmark & 34.4 & 35.2 & 25.7 & 23.2 & 21.1 \\
            \midrule
            \multirow{2}{*}{LoRA Fine-tuned} & VideoLLaMA2 & \checkmark & & 67.9 & 66.0 & 66.5 & \textbf{79.8} & \textbf{76.7} \\
            & \ours & \checkmark & \checkmark & \textbf{77.6} & \textbf{66.6} & \textbf{67.2} & 78.5 & 75.5\\
            \bottomrule
        \end{tabular}
        }
        \vspace{-8pt}        
        \caption{Performance of \ours when a pre-trained VideoLLaMA2 is used. \ours outperforms fine-tuned VideoLLaMA2 on benchmarks that focus more on spatial understanding. Notably, the performance gap on CLEVRER-MC is significant.}
        \vspace{-12pt}
    \label{tab:videollama2}
    \end{table*}
    
\subsection{Influence of each modality}
\label{sec:ablation_modality}

In \Cref{sec:compare_adapter}, the language-based representation is proved to be a more effective bounding box adapter.
In this section, we train \ours with the language-based box adapter and incorporate visual embeddings, video frame captions, and object bounding boxes in one model.
We also ablate the combinations of modalities to break down their contributions to performance.
The results are shown in \Cref{tab:ablation_modality}.

On CLEVRER-MC and Perception Test, the bounding-box-only model outperforms the video-only and caption-only models. And the caption-and-box model outperforms the caption-only model by a large margin on STAR.
This indicates the importance of object-centric information on these benchmarks. Adding boxes to video-only baseline leads to consistent improvements, indicating that the visual embeddings alone are not sufficient to encode objects.

The most significant improvement achieved by bounding boxes is on CLEVRER-MC, whose questions focus on object motion and collision.
Our qualitative results in \Cref{app:qualitative} show that our model can easily identify moving objects from the bounding boxes, which is difficult to determine from the frame captions.
We further break down the accuracy of different question types on CLEVRER-MC in \Cref{tab:breakdown_clevrer}.
We find that the improvement on object existence, moving count, and moving attribute is large, but is less significant for moving direction and counterfactual inference.
While counterfactual inference requires high-level reasoning, the moving direction of an object should be easily inferred from its bounding boxes.
However, we find that the training data we use does not include questions about direction.
This highlights that the learned understanding capability on symbolic representation cannot be perfectly generalized to all the tasks that are not involved during training.

\begin{figure*}[t]
    \centering
    \includegraphics[width=.9\linewidth]{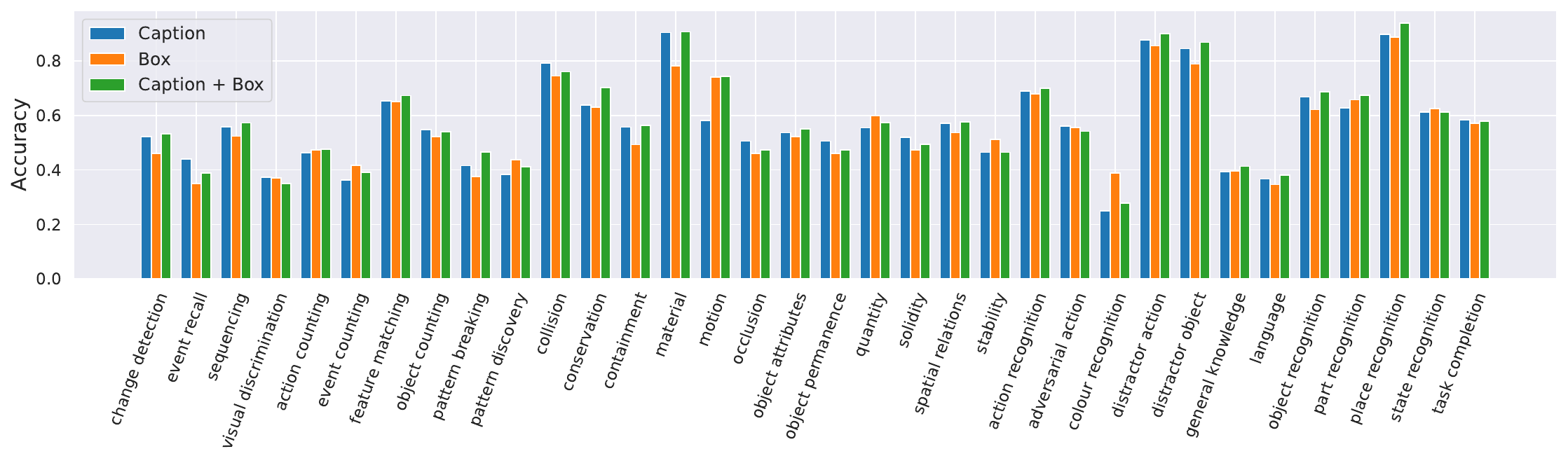}
    \vspace{-10pt}
    \caption{Accuracy of difference types of questions on Perception Test. Bounding boxes bring notable improvement on motion questions.}
    \label{fig:pt_breakdown}
    \vspace{-4pt}
\end{figure*}

The accuracy breakdown on Perception Test in \Cref{fig:pt_breakdown} shows substantial improvement on motion questions. The qualitative example in \Cref{fig:q_ptest} shows that the model can infer the spatial relation of the objects based on bounding boxes. Meanwhile, captions only miss the precise location of the truck, which is critical to answer the question.
More qualitative examples are in \Cref{app:qualitative}.

On NExT-QA and IntentQA, the box-only model cannot achieve better performance than the caption-only model and the video-only model.
As discussed in \Cref{app:qualitative}, these benchmarks focus on human actions and causal reasoning of events, which are difficult to be represented by object bounding boxes alone.
This shows that spatiotemporal object information is not equally important on all benchmarks.

Finally, while our caption-and-box models can always beat or be on par with caption-only and box-only models, integrating visual embedding does not improve the performance over the caption-and-box models on any benchmark.
This result is in line with Vamos \cite{wang2023vamos}, which highlights the difficulty in training effective embedding projectors for distributed representations with limited data.

    \begin{table*}[t]
    \centering
    \scalebox{0.83}{
        \begin{tabular}{c|c|cccccc}
            \toprule
            Models & Size & CLEVRER-MC & Perception Test & STAR & NExT-QA & IntentQA  \\
            \midrule
            \textcolor{gray}{w/ pre-trained visual adapter} \\
            \midrule
            LLaVA-Next-Video-DPO \cite{zhang2024llavanextvideo} & 7B & 38.4$^*$$^\dagger$ & 49.3$^*$ & - & - & - \\
            VideoLLaMA2 \cite{damonlpsg2024videollama2} & 7B & 45.6$^{\dagger}$ & 51.4$^*$ & 57.1$^*$$^{\dagger}$ & 74.1$^{\dagger}$ & 73.8$^*$$^{\dagger}$ \\ 
            SeViLA \cite{yu2023self} & 3B & - & 62.0 &  64.9 & 73.8 & - \\
            ViLA \cite{wang2024vila} & 3B & - & - & 67.1 & 75.6 & - \\
            \ours (VideoLLaMA2) & 7B & \textbf{77.6} & \textbf{66.6} & \textbf{67.2} & \textbf{78.5} & \textbf{75.5}\\
            \midrule
            \textcolor{gray}{w/o pre-trained visual adapter} \\
            \midrule
            Vamos \cite{wang2023vamos} & 8B &- & 62.3 & 63.7 & \textbf{77.3} & 74.2 \\
            \ours (LLaMA3) & 8B & \textbf{75.5} & \textbf{65.7} & \textbf{64.4} & 76.6 & \textbf{75.6} \\
            \bottomrule
        \end{tabular}}
    \vspace{-8pt}
    \caption{Comparison with existing MLLMs on five video QA benchmarks. Equipped with detected object bounding boxes, \ours achieves consistent improvements over baseline methods without explicit object representations, when starting from both an MLLM with pre-trained visual adapters, or an LLM that takes video captions as inputs. $^*$: Zero-shot generalization performance. $^\dagger$: Reproduced by us.} 
    \label{tab:main_results}
    \vspace{-10pt}
    \end{table*}

\subsection{Boosting pre-trained MLLMs with objects}\

We further study whether object representation can boost the performance of pre-trained MLLMs, which may already implicitly encode object information via their visual adapters.
We develop \ours from VideoLLaMA2 by including both the regular visual inputs and the language-represented object bounding boxes in the inputs.
\Cref{tab:videollama2} shows that \ours with pre-trianed VideoLLaMA2 backbone cannot understand the bounding boxes in a zero-shot manner.
However, after LoRA fine-tuning the model with video and boxes as inputs on the target benchmarks, \ours outperforms VideoLLaMA2 fine-tuned with only video inputs on CLEVRER-MC, Perception Test, and STAR.
These results show that the object bounding boxes provide additional information over what VideoLLaMA2 can get from distributed visual embeddings.
Perhaps not surprisingly, the relative gains are smaller compared to Table~\ref{tab:ablation_modality} as object information has already been partially integrated via visual adapters.

\subsection{Comparison with existing MLLMs}

In \Cref{tab:main_results}, we compare the performance of \ours with existing MLLMs, including models with large-scale pre-trained visual adapter \cite{damonlpsg2024videollama2,zhang2024llavanextvideo,yu2023self,wang2024vila} and models without it \cite{wang2023vamos}.
With object bounding boxes, \ours consistently outperforms other MLLMs in both settings.
The gaps are significant on CLEVRER-MC and Perception Test, which reveals the weakness of existing MLLMs in understanding spatiotemporal object configurations.

    \begin{table}[t]
      \centering
      \hspace{-5pt}
        \centering
        \scalebox{0.9}{
            \begin{tabular}{c|ccccc}
                \toprule
                Modality & Video & Caption & Pose & V + P & C + P \\
                \cmidrule(r){1-6}
                Accuracy & 66.6 & 60.2 & 63.5 & 68.6 & \textbf{69.4} \\ 
                \bottomrule
            \end{tabular}}
        \vspace{-8pt}
        \caption{Evaluation results on BABEL-QA~\cite{endo2023humanmotionqa}. Human poses bring improvements over video embeddings and frame captions.}
        \label{tab:humanmotionqa}
      \vspace{-15pt}
    \end{table}

\subsection{Generalize to richer symbolic representations}
Language-based representation is also a natural way to incorporate structured visual representations other than object bounding boxes.
For example, human pose can be represented by a list of keypoint names and coordinates in texts.

To demonstrate the effectiveness of \ours with human pose, we use BABEL-QA~\cite{endo2023humanmotionqa}, a human motion QA benchmark that focuses on human activity understanding.
Each example in BABEL-QA has a sequence of 3D human keypoints over time and asks a question about their actions.
In \ours, we sample six frames from the sequence and represent the poses in text, with all coordinates normalized to integers within $[0,1000]$, for example,
    \begin{tcolorbox}[colback=green!5!white, colframe=green!50!black]
    Frame 0: pelvis [500 500 542] left hip [497 499 538] right hip [502 499 538] spine [499 498 548] ...
    \end{tcolorbox}

\noindent We render the poses as videos and employ CLIP for visual embeddings and GPT-4o for captions.
\Cref{tab:humanmotionqa} demonstrates the effectiveness of human poses, and that \ours generalizes to richer structured visual representations.

\section{Conclusion}

We investigate how can objects help video-language understanding in the context of multimodal large language models. We demonstrate the effectiveness of  object-centric representations extracted by off-the-shelf computer vision algorithms.
Unlike distributed visual representations such as CLIP, object-centric representations can be integrated into MLLMs in a data-efficient manner, such as by rendering as plain text. We introduce \ours that utilizes object-centric representations via bounding box adapters, and demonstrate that \ours achieves competitive performance over approaches that utilize only visual embeddings or captions across six video QA benchmarks. We also observe that \ours generalizes to richer structured visual representations, such as human pose on the BABEL-QA benchmark.
We believe our observations highlight the importance of explicitly integrating computer vision models into MLLMs via language-based, or other data-efficient interfaces, making vision a first-class citizen for vision-language models again. 

\noindent\textbf{Acknowledgments:} This work is supported by the Global Research Outreach program of Samsung. Our research was conducted using computational resources at the Center for Computation and Visualization at Brown University. We appreciate valuable feedback from Calvin Luo, Tian Yun, Yuan Zang, and Zilai Zeng.

{
    \small
    \bibliographystyle{ieeenat_fullname}
    \bibliography{main}
}
\clearpage
\setcounter{page}{1}
\setcounter{table}{0}
\renewcommand{\thetable}{A\arabic{table}}
\setcounter{figure}{0}
\renewcommand{\thefigure}{A\arabic{figure}}
\maketitlesupplementary

\appendix

We elaborate on the implementation details of \ours in \Cref{app:detail}.
In \Cref{app:ablation}, we explore the design choices for the bounding box projector, the effectiveness of object bounding boxes compared to object labels, the modality fusion strategy, and bounding box sampling rate.
In \Cref{app:nextqa_and_intentqa}, we conduct experiments on NExT-QA and IntentQA to show that spatial cues play a minor role on these two benchmarks.
\Cref{app:box_quality} demonstrates the quality of the detected object bounding boxes and investigates how it affects the performance of \ours.
Qualitative results of our method are presented in \Cref{app:qualitative}.
Finally, \Cref{app:unsuccessful} summarizes a few unsuccessful attempts.

\section{Implementation Details}
\label{app:detail}

\subsection{Object detection and tracking}
In the object detection and tracking process, the video keyframes are sampled at 1 FPS. SAM 2 tracking is performed at the original frame rate of each video.
We use the pre-trained YOLO-World checkpoint \texttt{YOLO-World-v2-L\_CLIP-Large\_800}.
The employed SAM 2 checkpoint is \texttt{sam2.1\_hiera\_large}.

All the benchmarks (or their source datasets) in our experiments provide either manually annotated or algorithm-detected object bounding boxes.
To adapt YOLO-World to the benchmarks, we fine-tune it on the training set of each benchmark individually.
Fine-tuning is performed with a learning rate of $2e-4$, a weight decay of $0.05$, and a batch size of $64$.
The number of training images, the number of training epochs, and the score thresholds used during inference are listed in \Cref{tab:tracking_detail}.

\subsection{Downsampling Rates of Bounding Boxes}
\label{app:downsample}

\begin{figure}[t]
        \centering
            \scalebox{1}{
            \includegraphics[width=0.94\linewidth]{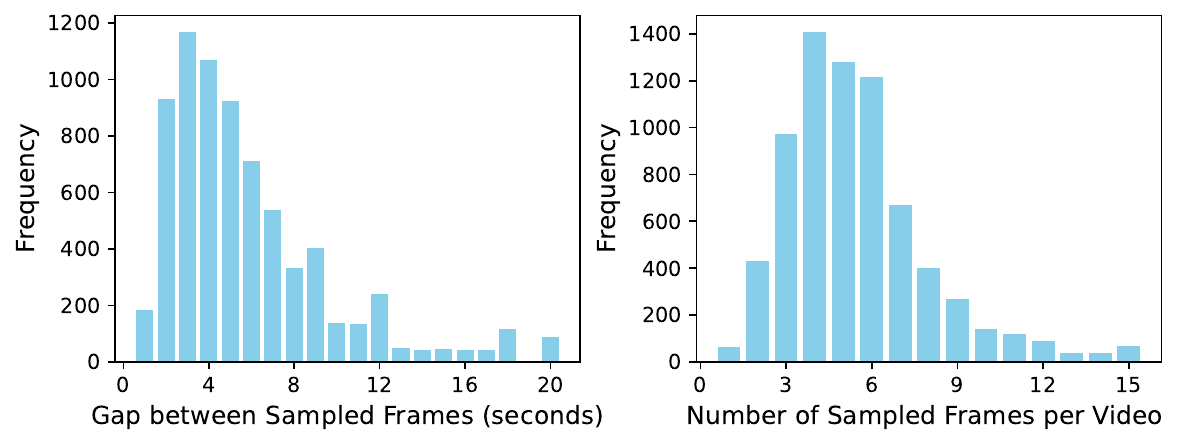}
            }
        {\small (a) Perception Test}
            \scalebox{1}{
            \includegraphics[width=0.94\linewidth]{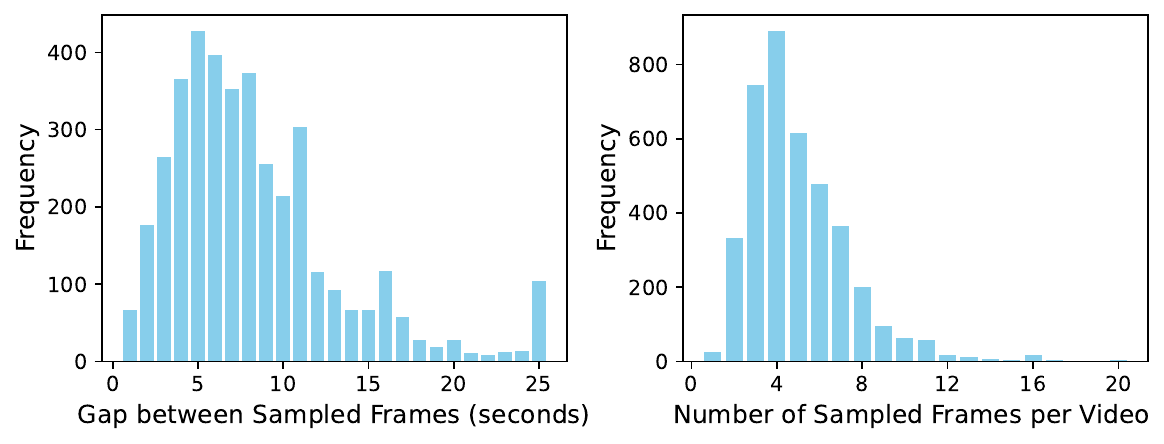}
            }
        {\small (b) STAR}
            \scalebox{1}{
            \includegraphics[width=0.94\linewidth]{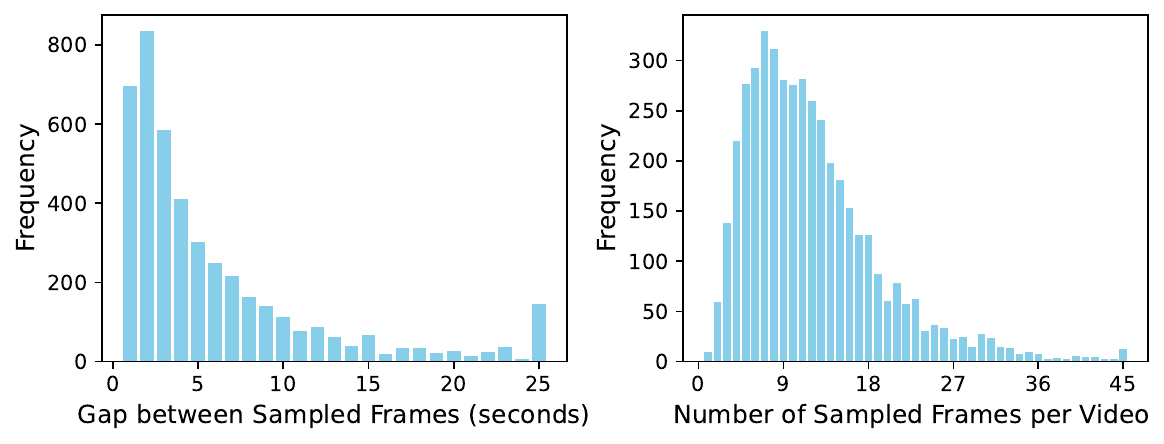}
            }
        {\small (c) NExT-QA \& IntentQA}
        \vspace{-7pt}
            \caption{Distributions of the bounding box sampling rates for Perception Test \cite{patraucean2023perception}, STAR \cite{wu2021star_situated_reasoning}, NExT-QA \cite{xiao2021next}, and IntentQA \cite{Li_2023_ICCV}. We show the gaps between the sampled frames and the numbers of the sampled frames. The last bin includes all elements greater than or equal to the corresponding x-coordinate value. IntentQA shares the same distribution as NExT-QA because it is sourced from NExT-QA. 
            }
            \label{fig:sampling_rate}
            \vspace{-5pt}
    \end{figure}

    As described in \Cref{sec:implementation}, we uniformly and temporally downsample the bounding box sequences to reduce the total number of input tokens.
    As the videos in CLEVRER-MC \cite{CLEVRER2020ICLR,li2023mvbench} are 5 seconds in length, we sample one frame every one second, resulting in 6 sampled frames per video.
    For other benchmarks, the videos have varying lengths and numbers of objects.
    Therefore, we assign a separate sampling rate for each video.
    Specifically, we use binary search to find the maximal sampling rate for each video such that the number of bounding box tokens after downsampling is less than 1,000.
    We show the distributions of the sampling rates and the resulting numbers of frames in \Cref{fig:sampling_rate}.

    \begin{table}[t]
    \centering
    \scalebox{0.76}{
        \begin{tabular}{c|ccc}
            \toprule
            Benchmark & \#training images & \#epochs & score threshold \\
            \cmidrule(r){1-4}
             CLEVRER-MC & 60 k & 7 & $1e-3$ \\
             Perception Test & 46 k & 50 & 0.35 \\
             STAR & 106 k & 20 & 0.2 \\
             NExT-QA \& IntentQA & 151 k & 10 & 0.4 \\
            \bottomrule
        \end{tabular}
        }
    \vspace{-5pt}
    \caption{Hyperparameters in YOLO-World fine-tuing and inference across different benchmarks.}
    \label{tab:tracking_detail}
    \end{table}

\subsection{ObjectMLLM training}

When fine-tuning LLaMA3-8B with LLaMA-Adapter \cite{zhang2023llamaadapter}, we use a batch size of 64.
The learning rate is linearly warmed up to $0.0225$ in the first $20\%$ steps, after which cosine learning rate annealing is applied.
The same learning rate is applied to the LLaMA-Adapter weights, bounding box embedding projector, and visual embedding projector.

When fine-tuning VideoLLaMA2 with LoRA \cite{hu2022lora}, we use a LoRA rank of 128 and a batch size of 128.
The learning rate is linearly warmed up to $2e-5$ in the first $3\%$ steps, after which cosine learning rate annealing is applied.
The same learning rate is applied to the LoRA weight and visual embedding projector.
The pre-trained checkpoint used is \texttt{VideoLLaMA2-7B-16F}.

In both settings, the model is trained for 1 epoch on CLEVRER, 5 epochs on NExT-QA and STAR, 10 epochs on Perception Test and IntentQA, and 50 epochs on BABEL-QA.
The vision encoder is kept frozen in all experiments.

\section{Ablation Studies}
\label{app:ablation}

\subsection{Embedding projector}

    We show that the language-based box adapter is always more performant than the embedding projector in \Cref{sec:compare_adapter}.
    However, it is possible that the poor performance of the embedding projector is due to its design.
    In this section, we explore a few design choices of the embedding projector, including initialization, number of layers, and number of resulting tokens.
    We find that the embedding projector is inferior to the language-based representation across all design choices.

    \noindent\textbf{Projector initialization.}
    When training a projector between a novel modality and the LLM backbone, previous works (\eg LLaVA \cite{liu2023llava} and Vamos \cite{wang2023vamos}) use the default initialization for linear layers (the Kaiming uniform distribution in PyTorch).
    However, we find that the default random initialization significantly impedes the training of the bounding box embedding projector.
    As shown in \Cref{tab:init}, we find that initializing the linear layer weights to zero can facilitate the learning of embedding projector.
    We hypothesize that the default initialization would project the bounding boxes outside the LLM word embedding space, confusing the LLM backbone at the beginning of training.
    In contrast, a zero-initialized linear layer can map every bounding box to a zero vector, which is extremely close to the special tokens in the LLM vocabulary.
    This can prevent the bounding boxes from corrupting LLM behavior.

    \begin{table}[t]
    \centering
    \scalebox{0.75}{
        \begin{tabular}{c|c|cc}
            \toprule
            Box adapter & Initialization & CLEVRER-MC & Perception Test  \\
            \cmidrule(r){1-4}
            \multirow{2}{*}{Embedding Projector} & Random & 42.8 & 57.6 \\
             & Zero & \textbf{64.9} & \textbf{60.1} \\
            \bottomrule
        \end{tabular}
        }
    \vspace{-5pt}
    \caption{Ablation on the initialization of the box embedding projector. The defulat random initialization in PyTorch is Kaiming uniform distribution. We find that zero-initialized linear layer as the box embedding projector always yields better performance than the default initialization.}
    \label{tab:init}
    \end{table}
    
    \noindent\textbf{Number of projector layers.}
    Instead of a single linear layer, we explore using a multilayer perceptron (MLP) as the bounding box projector.
    In this experiment, we set the number of hidden units in each layer to match the dimension of word embeddings of the LLM (4,096 for LLaMA3-8B).
    We use GeLU as the activation function.
    As \Cref{tab:n_layers} suggests, increasing the number of MLP layers yields only marginal performance gains for the embedding projector method.
    More importantly, it is still dominated by the language-based representation approach.

    \begin{table}[t]
    \centering
    \scalebox{0.80}{
        \begin{tabular}{c|c|cc}
            \toprule
            Box adapter & \#Layers & CLEVRER-MC & Perception Test  \\
            \cmidrule(r){1-4}
            & 1 & 64.9 & 60.1 \\
            Embedding Projector & 2 & 65.0 & 59.7 \\
            & 3 & 65.0 & 60.2 \\
            \midrule
            Language-based  & \multirow{2}{*}{-} & \multirow{2}{*}{\textbf{77.6}}  & \multirow{2}{*}{\textbf{63.5}} \\
            Representation & \\
            \bottomrule
        \end{tabular}
        }
    \vspace{-5pt}
    \caption{Ablation on the number of layers of the box embedding projector. Enlarging the number of MLP layers does not bring significant improvement. And they are outperformed by the language-based representation box adapter.}
    \label{tab:n_layers}
    \end{table}
    
    \noindent\textbf{Number of resulting tokens.}
    While the embedding projector maps each bounding box into only one token, the language-based representation uses nine tokens to describe one bounding box (four numbers, three spaces, and two square brackets).
    One might argue that the expressiveness of the bounding box embedding projector is limited by the number of tokens.
    To address this concern, we experiment with bounding box projectors that map each bounding box into nine tokens rather than one token.
    The results are in \Cref{tab:n_tokens}.
    We find that increasing the number of resulting tokens per bounding box cannot improve the performance of the embedding projector method.

    \begin{table}[t]
    \centering
    \scalebox{0.78}{
        \begin{tabular}{c|c|c}
            \toprule
            Box adapter & \#Tokens per box & CLEVRER-MC \\
            \cmidrule(r){1-3}
            \multirow{2}{*}{Embedding Projector} & 1 & 64.9 \\
             & 9 & 63.4 \\
            \midrule
            Language-based  & \multirow{2}{*}{9} & \multirow{2}{*}{\textbf{77.6}} \\
            Representation & & \\
            \bottomrule
        \end{tabular}
        }
    \vspace{-5pt}
    \caption{Ablation on the number of resulting tokens in the embedding projector method. Increasing the number of tokens to be the same as that in the language-based representation method degrades the performance.}
    \label{tab:n_tokens}
    \end{table}

    \begin{table*}[t]
    \centering
        \begin{tabular}{ccc|ccccc}
            \toprule
            Video & Caption & Box & CLEVRER-MC & Perception Test & STAR & NExT-QA & IntentQA  \\
            \cmidrule(r){1-8}
            \checkmark & & & 40.3 & 59.6 & 59.7 & 70.7 & 68.2 \\ 
            & \checkmark & & 47.8 & 62.4 & 60.1 & 76.6 & 75.7 \\ 
            & & \checkmark & 77.6 & 63.5 & 59.1 & 63.7 & 66.2 \\ 
            & \checkmark & \checkmark & 75.5\blue{(75.8)} & 65.7\blue{(64.1)} & 64.4\blue{(63.7)} & 76.6\blue{(77.2)} & 75.6\blue{(75.4)} \\ 
            \checkmark & \checkmark & \checkmark &  75.4\blue{(29.5)} & 63.9\blue{(33.8)}  & 62.9\blue{(62.8)} & 76.2\blue{(75.8)} & 75.0\blue{(73.4)} \\ 
            \bottomrule
        \end{tabular}
    \vspace{-5pt}
    \caption{Ablation on modality fusion strategy. \blue{Blue ones are the results of jointly training on all the modalities at once}, while the others are trained in a modality-by-modality manner. The modality-by-modality fusion strategy can outperform the jointly training in most cases.
    And the joint training is sometimes unstable when the video inputs are involved.
    }
    \label{tab:ablation_fusion}
    \end{table*}

    \begin{table}[t]
    \centering
        \scalebox{0.88}{
        \begin{tabular}{c|ccc}
            \toprule
            Input & CLEVRER-MC & Perception Test & STAR \\
            \midrule
            Obj. label & 59.8 & 60.0 & 58.8 \\
            Obj. label + box & \textbf{77.6} & \textbf{63.5} & \textbf{59.1} \\
            \bottomrule
        \end{tabular}
        }
    \vspace{-5pt}
    \caption{Ablation on the bounding boxes versus object labels. The model indeed utilizes the boxes other than the object labels.}
    \label{tab:object_label_ablation}
    \end{table}

\subsection{Are object labels sufficient on their own?}
    As shown in \Cref{fig:bbox_template}, object labels (i.e., object names) are also provided when we format the bounding boxes.
    If the object labels are hidden, the bounding boxes themselves convey much less information because the object associated with each box is unknown.
    Object labels can provide important information to the model; for example, color recognition in \Cref{fig:pt_breakdown} is improved, which is definitely not inferable from unannotated bounding boxes alone.
    We raise the question: are object labels sufficient on their own, or does the model still derive benefits from bounding box information?
    
    To answer this question, we train a model with object labels provided but bounding boxes hidden.
    In \Cref{tab:object_label_ablation}, we find that the model consistently performs better when the object boxes are also provided, indicating that the model makes effective use of object bounding boxes.
    However, the differences are more notable on CLEVRER-MC and Perception Test than on STAR, indicating the improvement made by observing bounding boxes on STAR is mainly attributed to the revealed object labels. 
    This is reasonable because the questions in STAR are annotated based off the scene graphs.
    It also highlights that emphasizing the question-related objects instead of describing the scene at a high level through video frame captions helps more on video question answering.

\subsection{Modality fusion strategy}
    \Cref{sec:finetune_strategy} mentions that we fuse the modalities (captions, bounding boxes, and videos) in a modality-by-modality approach instead of joint training at once.
    In \Cref{tab:ablation_fusion}, we compare the two fusion strategies.
    We find that the modality-by-modality method outperforms joint training in the majority of cases.
    In addition, the joint training approach is sometimes unstable.
    It leads to extremely low performance on CLEVRER-MC and Perception Test when using all three input modalities.
    Lastly, we find that both fusion methods are ineffective at leveraging the visual embeddings.
    We therefore urge the development of new multimodal fusion strategies that can make visual inputs valuable.

    \begin{table}[t]
        \begin{subtable}{.5\linewidth}
          \centering
            \scalebox{0.59}{
            \begin{tabular}{c|cccc}
                \toprule
                FPS & 0.25 & 0.5 & 1 & 2 \\
                \midrule
                Accuracy & 74.0 & 73.9 & 77.6 & 78.0 \\
                \bottomrule
            \end{tabular}
            }
            \caption{CLEVRER-MC}
        \end{subtable}%
        \begin{subtable}{.5\linewidth}
          \centering
          \scalebox{0.59}{
            \begin{tabular}{c|cccc}
                \toprule
                Max Seq. Len. & 600 & 800 & 1000 & 1200 \\
                \midrule
                Avg. FPS & 0.13 & 0.20 & 0.26 & 0.31 \\
                \midrule
                Accuracy & 62.7 & 63.0 & 63.5 & 63.7 \\
                \bottomrule
            \end{tabular}
            }
            \caption{Perception Test}
        \end{subtable}
        \vspace{-20pt}
        \caption{Ablation on bounding box temporal sampling rate. Higher sampling rate usually leads to higher accuracy.}
        \label{tab:sampling_rate}
        \vspace{-10pt}
    \end{table}

\subsection{Temporal sampling rate of bounding boxes}

We conduct an ablation study on the bounding box sampling rate on CLEVRER-MC and Perception Test in Table~\ref{tab:sampling_rate}.
On CLEVRER-MC, all the videos have the same duration. Therefore, we directly adjust the bounding box sampling rate from 0.25 FPS to 2 FPS.
On Perception Test, we vary the maximum box token length mentioned in \Cref{sec:implementation} and report the corresponding average frame rates.
As the results show, a higher sampling rate generally yields higher accuracy, indicating a trade-off between performance and efficiency.

    \begin{figure}[t]
        \centering
        \includegraphics[width=\linewidth]{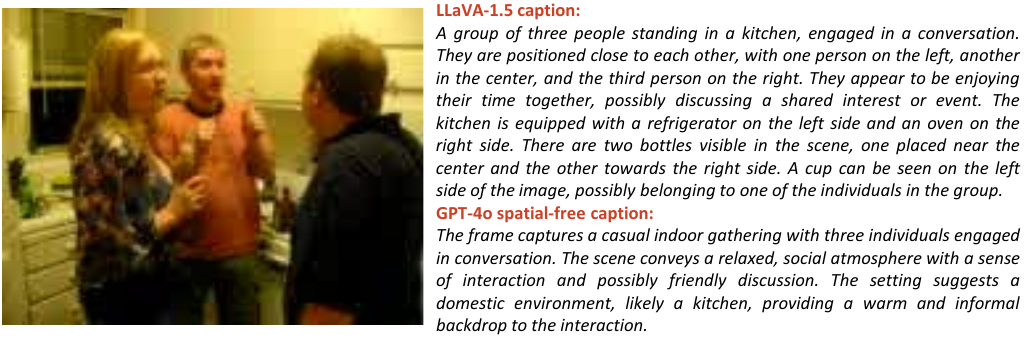}
        \vspace{-22pt}
        \caption{LLaVA-1.5 caption vs. GPT-4o spatial-free caption.}
        \label{fig:gpt4o}
        \vspace{-10pt}
    \end{figure}
    
    \begin{table}[t]
    \centering
    \scalebox{0.9}{
        \begin{tabular}{cc|cc}
            \toprule
            Caption & Box & NExT-QA & IntentQA  \\
            \cmidrule(r){1-4}
            LLaVA-1.5 & & 76.6 & 75.7 \\
            LLaVA-1.5 & \checkmark & 76.6 & 75.6 \\
            GPT-4o Spatial-free & & 76.1 & 73.6 \\
            GPT-4o Spatial-free & \checkmark & 76.3 & 75.6 \\
            \bottomrule
        \end{tabular}}
    \vspace{-8pt}
    \caption{Performance with spatial-free GPT-4o captions.}
    \label{tab:caption}
    \vspace{-12pt}
    \end{table}

    \begin{table*}[t]
    \centering
    \scalebox{0.9}{
        \begin{tabular}{c|ccccc}
            \toprule
            Box & CLEVRER-MC & Perception Test & STAR & NExT-QA & IntentQA  \\
            \cmidrule(r){1-6}
            Model-tracked & \textbf{77.6} & 63.5 & 59.1 & 63.7 & \textbf{66.2} \\ 
            Annotation & 74.9 & \textbf{66.8} & \textbf{78.9} & \textbf{65.5} & 65.3 \\ 
            \bottomrule
        \end{tabular}}
    \vspace{-5pt}
    \caption{Model performance with object bounding boxes tracked by computer vision models or with those annotated by the benchmarks. Bounding box annotations in CLEVRER-MC, NExT-QA, and IntentQA are also algorithm-detected. Boxes in Perception Test and STAR are manually annotated. The experiments are in the box-only setting.}
    \label{tab:box_quality}
    \vspace{-10pt}
    \end{table*}

\section{Spatial Information Ablation on NExT-QA and IntentQA}
\label{app:nextqa_and_intentqa}

In our experiments in \Cref{tab:ablation_modality} and \Cref{tab:videollama2}, incorporating object bounding boxes as additional information does not improve model performance on NExT-QA and IntentQA.
While these benchmarks focus on causal reasoning about events, the spatial information may not play a crucial role to answer the questions.
In this section, we further verify through experiments the limited usefulness of spatial information in these two benchmarks.

As \Cref{fig:gpt4o} shows, the video frame captions generated by LLaVA-1.5 include some spatial cues in the image.
To eliminate the influence of spatial cues, we feed each video frame to GPT-4o and ask it to generate a caption that does not involve any spatial information.
Specifically, we use the following prompt:
    \begin{tcolorbox}[colback=green!5!white, colframe=green!50!black]
    \textless image\textgreater\quad Faithfully describe the content of the above image, avoid mentioning specific objects and try your best to provide a scene-level, holistic summary. No mention of any spatial information.
    \end{tcolorbox}

\Cref{fig:gpt4o} shows an example of such spatial-free captions.
With the spatial-free captions, we repeat our experiments on NExT-QA and IntentQA, using \ours with LLaMA3 as the backbone.
The results are in \Cref{tab:caption}.
We find that removing the spatial cues only results in a 0.5\% drop in accuracy on NExT-QA and a 2.1\% drop on IntentQA.
This indicates that the questions in these two datasets are still highly answerable even without spatial information.
More importantly, incorporating object bounding boxes in addition to spatial-free captions can recover the performance, especially on IntentQA.
This again demonstrates our method's effectiveness in conveying spatial information to MLLMs.

\section{Quality of the Tracked Bounding Boxes}
\label{app:box_quality}
    
In this section, we assess the quality of the extracted object bounding boxes and whether it hinders the performance of our model.
Because the evaluation benchmarks themselves provide object bounding box annotations (either human-annotated or algorithm-detected), we can compare the boxes obtained by our workflow with them.

\begin{figure*}[t]
        \centering
            \scalebox{0.98}{
            \includegraphics[width=1\linewidth]{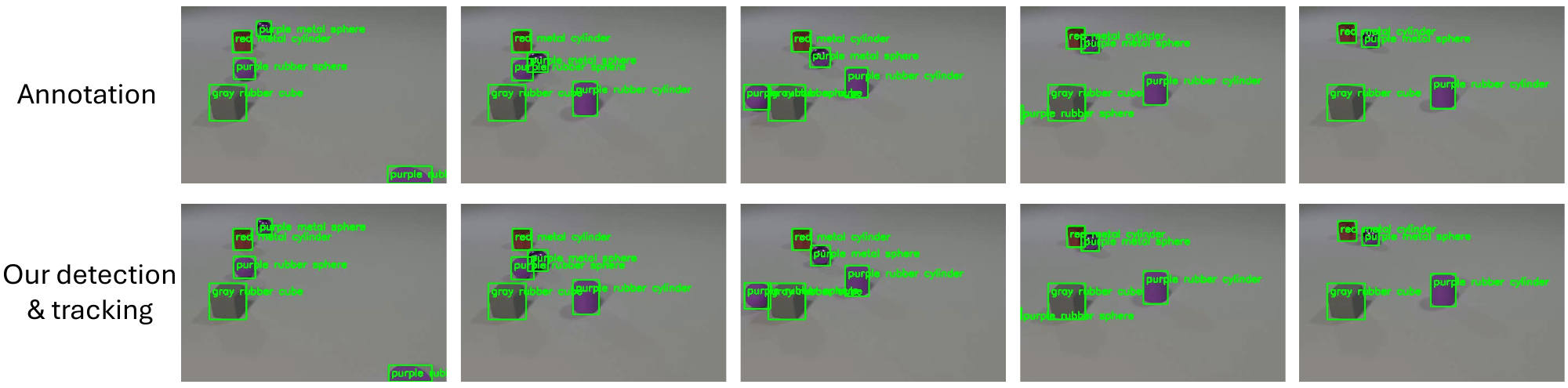}
            }
        {\small (a) CLEVRER-MC}
        \centering
            \scalebox{0.98}{
            \includegraphics[width=1\linewidth]{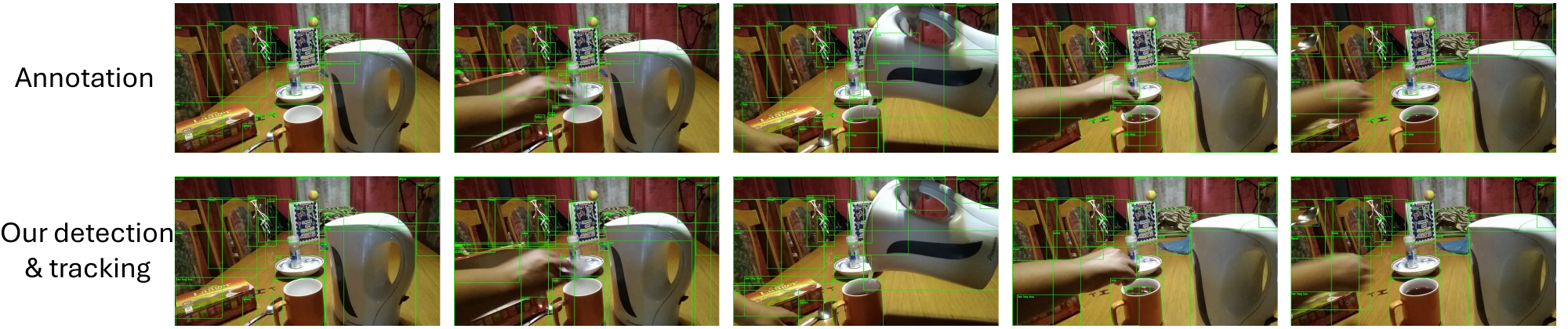}
            }
        {\small (b) Perception Test}
            \scalebox{0.98}{
            \includegraphics[width=1\linewidth]{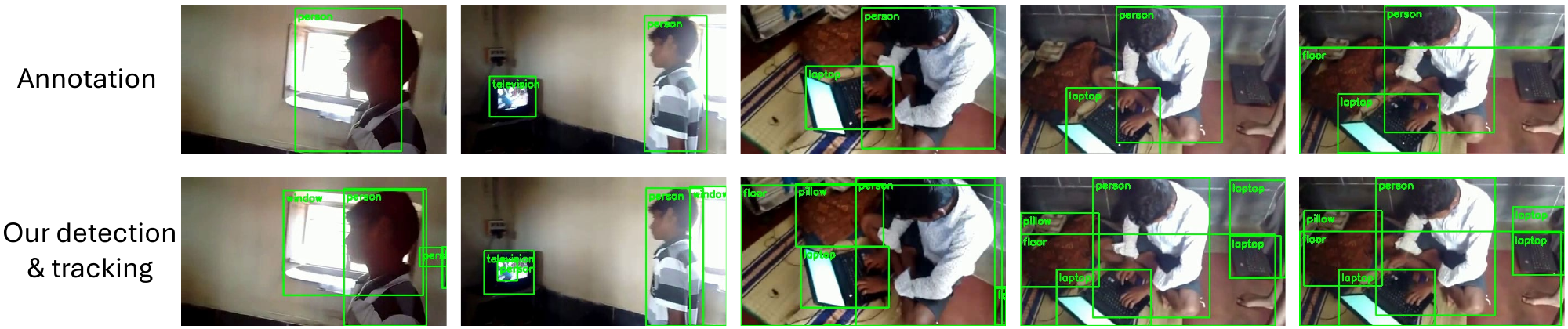}
            }
        {\small (c) STAR}
            \scalebox{0.98}{
            \includegraphics[width=1\linewidth]{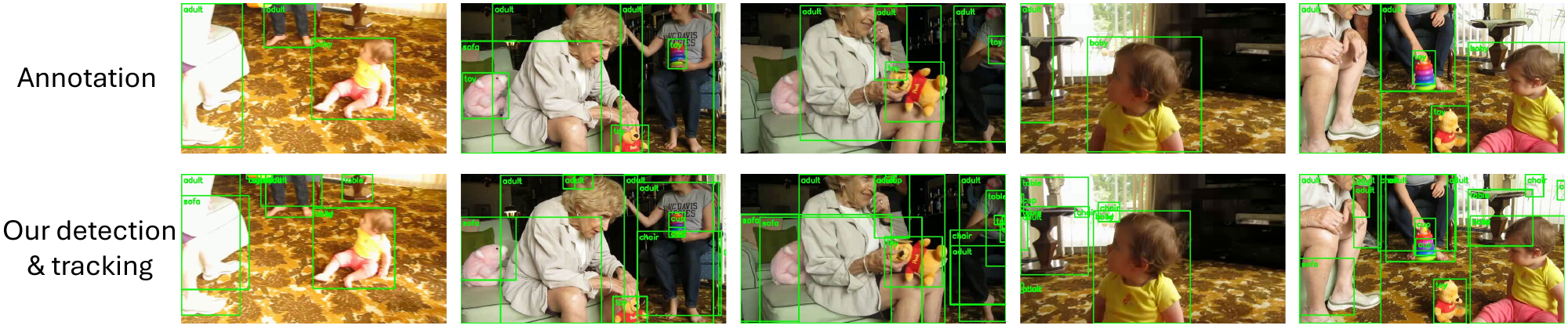}
            }
        {\small (d) NExT-QA \& IntentQA}
        \vspace{-7pt}
            \caption{Visualization of the object bounding boxes across different benchmarks. IntentQA shares the same video source as NExT-QA. Our tracked bounding boxes are nearly perfect on CLEVRER-MC, while they are also fairly accurate on realistic videos.
            }
            \label{fig:visualize_box}
            \vspace{-5pt}
    \end{figure*}

\Cref{fig:visualize_box} visualizes the tracked and annotated bounding boxes across different benchmarks.
All the examples are from the validation/test set of the benchmarks.
We find that the detection and tracking quality on synthetic videos (CLEVRER-MC) is highly accurate.
On the realistic videos (Perception Test, STAR, NExT-QA, and IntentQA), our employed tracking method can capture the main objects although some noise is present.

In \Cref{tab:box_quality}, we evaluate our model with the bounding box annotations as inputs.
When the annotations serve as inputs, the model is trained again with the annotated boxes before evaluation to mitigate train-test domain shift.
While the performance with model-tracked bounding boxes is $2\%\text{--}3\%$ worse than that with annotations on Perception Test and NExT-QA, it is even better than the annotations on CLEVRER-MC and IntentQA.
This is reasonable because the bounding box annotations provided in the CLEVRER and IntentQA benchmarks are also algorithm-detected and potentially noisier than ours.
In contrast, Perception Test and STAR both provide human-annotated object bounding boxes.
However, the performance gap on STAR is significantly larger than on Perception Test.
We notice that the questions in STAR are generated by functional programs based on annotated object relation graphs.
Because only objects of interest are annotated in STAR, using object annotations as input provides a strong prior that may bias the answers.
As the tracking quality on STAR (\Cref{fig:visualize_box}(c)) is fairly accurate, we hypothesize that the large performance gap is caused by the choices of objects of interest rather than the tracking precision.
How we can filter the objects of interest from a video remains an open and valuable research challenge.

\section{Qualitative Results}
\label{app:qualitative}

\begin{figure*}[t]
        \centering
        \includegraphics[width=\linewidth]{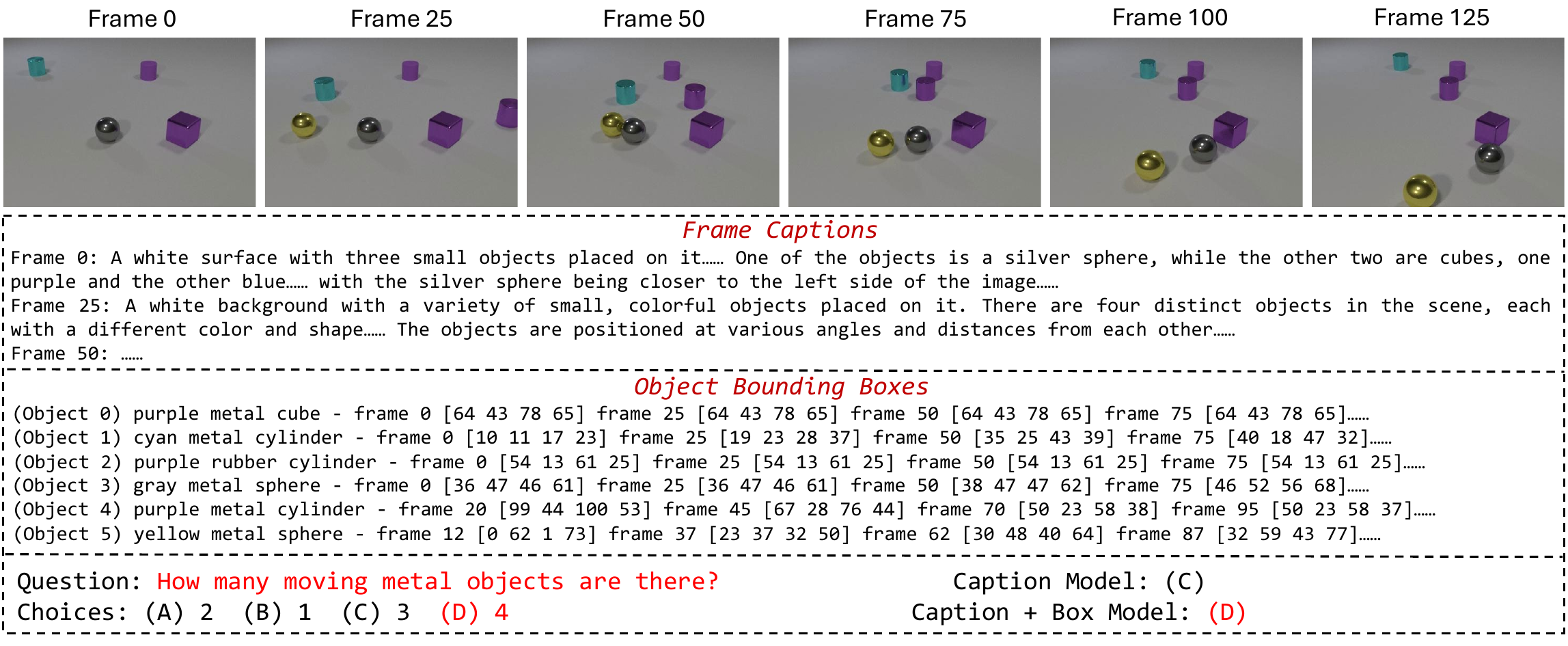}
        \vspace{-20pt}
        \caption{Qualitative example on CLEVRER-MC. The model can determine whether an object is moving based on its bounding boxes.}
        \label{fig:q_clevrer}
        \vspace{-5pt}
    \end{figure*}

    We show a qualitative result on CLEVRER-MC in \Cref{fig:q_clevrer}.
    The model can determine whether an object is moving based on its bounding boxes, which is an essential capability in this benchmark.
    
\begin{figure*}[t]
        \centering
        \includegraphics[width=\linewidth]{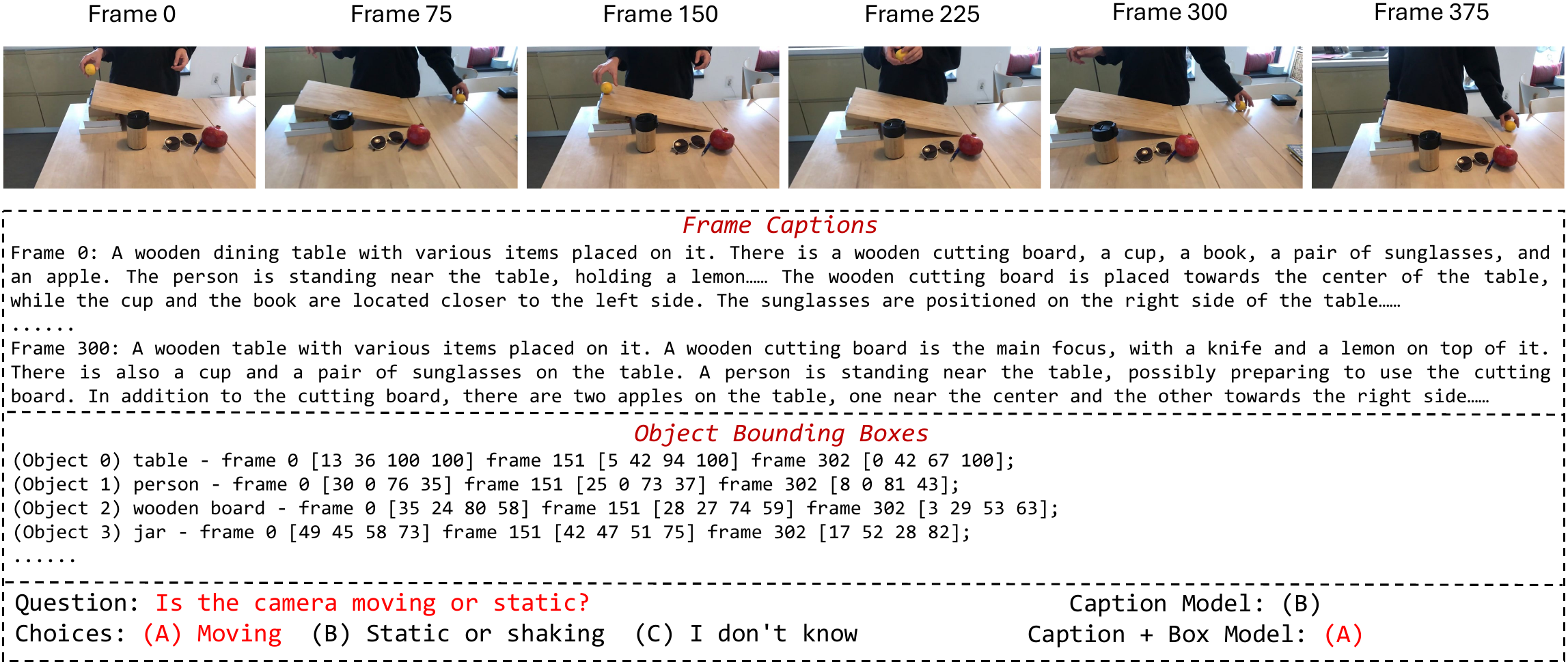}
        \vspace{-20pt}
        \caption{Qualitative example on Perception Test. The caption+box model can determine the motion of the camera from the changing object bounding boxes.}
        \label{fig:q_ptest_app1}
        \vspace{-5pt}
    \end{figure*}

\begin{figure*}[t]
        \centering
        \includegraphics[width=\linewidth]{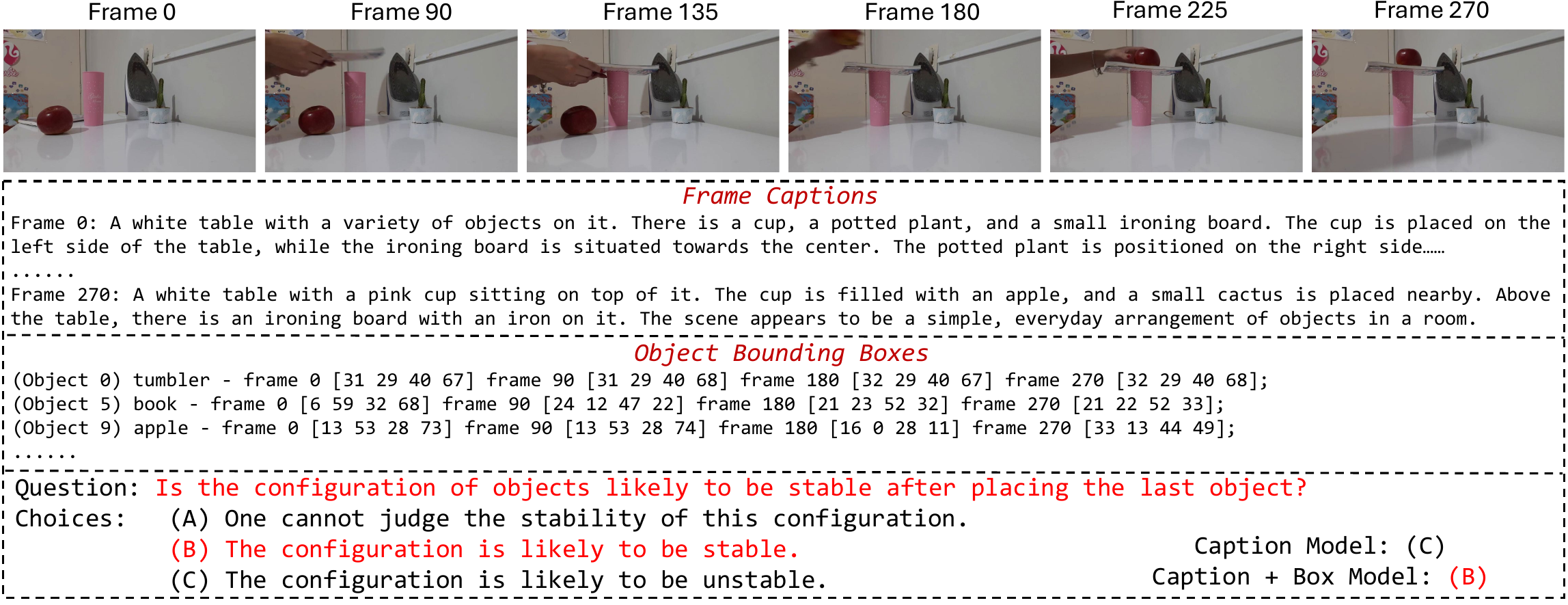}
        \vspace{-20pt}
        \caption{Qualitative example on Perception Test. The caption+box model can predict the stability of the object configuration because it is aware of the object locations.}
        \label{fig:q_ptest_app2}
        \vspace{-5pt}
    \end{figure*}

\begin{figure*}[t]
        \centering
        \includegraphics[width=\linewidth]{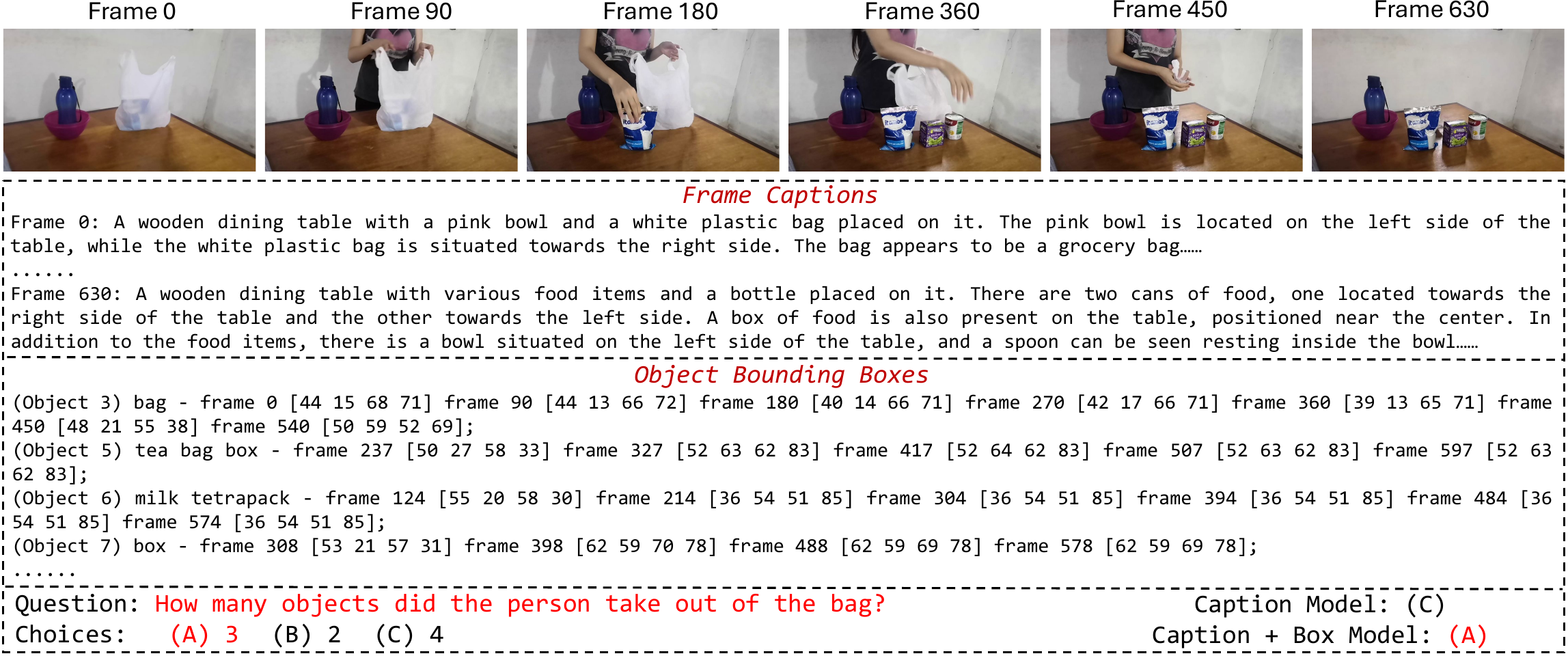}
        \vspace{-20pt}
        \caption{Qualitative example on Perception Test. The caption+box model can determine the number of objects taken out from the bag with the aid of object bounding boxes.}
        \label{fig:q_ptest_app3}
        \vspace{-5pt}
    \end{figure*}

\begin{figure*}[t]
        \centering
        \includegraphics[width=\linewidth]{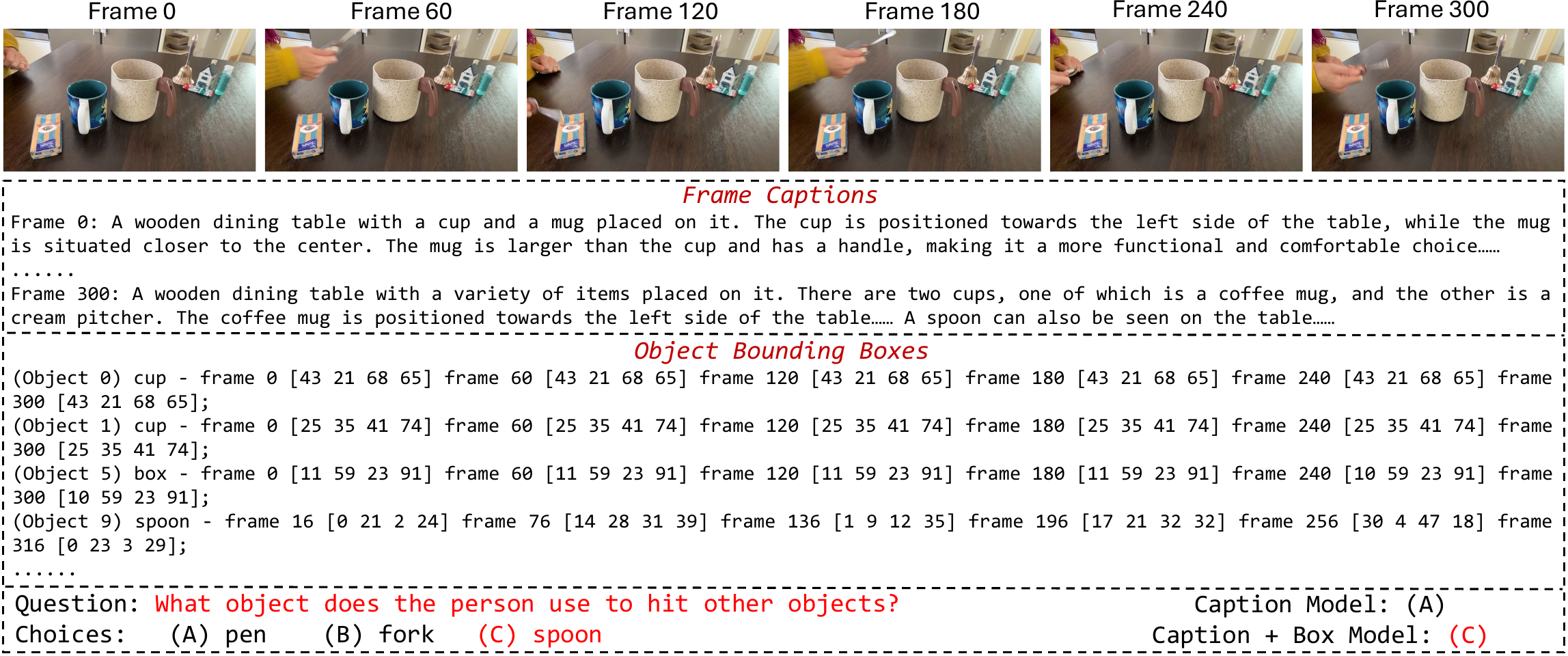}
        \vspace{-20pt}
        \caption{Qualitative example on Perception Test. From the boudning box cooridnates, the caption+box model can observe that the spoon is moved to hit the other objects.}
        \label{fig:q_ptest_app4}
        \vspace{-5pt}
    \end{figure*}

    In \Cref{fig:q_ptest_app1,fig:q_ptest_app2,fig:q_ptest_app3,fig:q_ptest_app4}, we show qualitative results from Perception Test.
    In these examples, model can determine the motion of cameras, stability of objection configurations, and the number of objects taken out from bags.
    These questions are not answerable for the caption-only model.

\begin{figure*}[t]
        \centering
        \includegraphics[width=\linewidth]{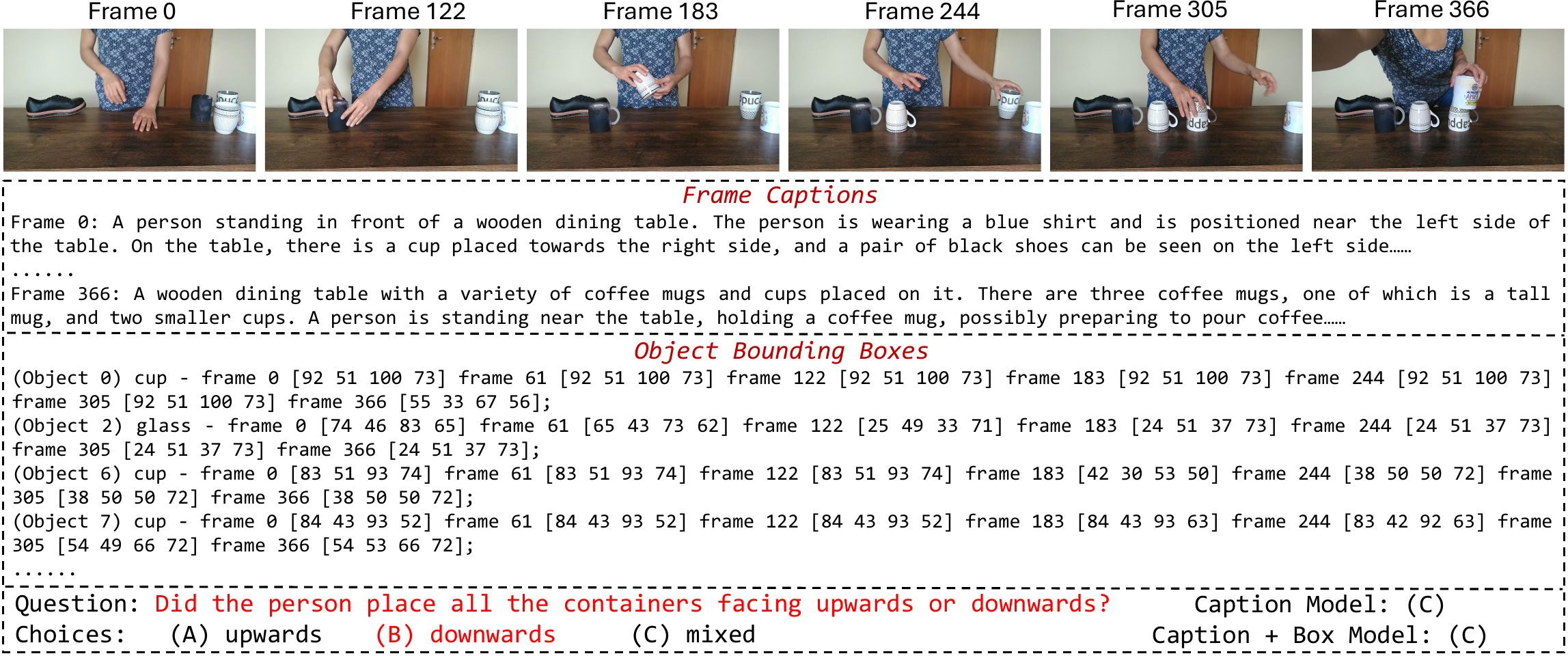}
        \vspace{-20pt}
        \caption{Failure case on Perception Test. The model cannot see the state of the mugs from either the captions or the bounding boxes. So it does not whether the mugs are upwards or downwards.}
        \label{fig:q_ptest_fail1}
        \vspace{-5pt}
    \end{figure*}

\begin{figure*}[t]
        \centering
        \includegraphics[width=\linewidth]{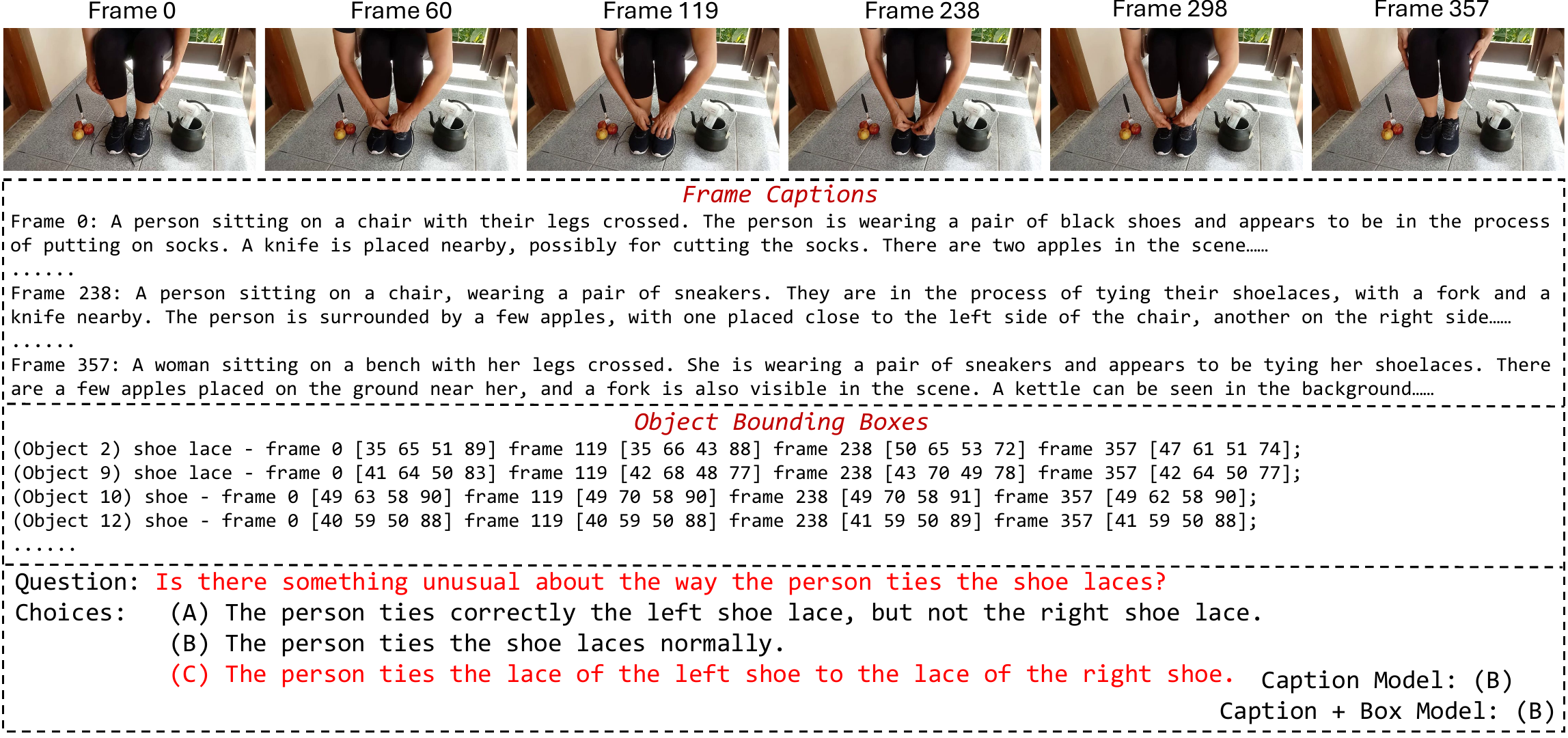}
        \vspace{-20pt}
        \caption{Failure case on Perception Test. Both the captions and the bounding boxes cannot tell if the shoe laces are tied normally. This suggests that our model has difficulty in recognizing the appearance of the objects.}
        \label{fig:q_ptest_fail2}
        \vspace{-5pt}
    \end{figure*}

    In \Cref{fig:q_ptest_fail1,fig:q_ptest_fail2}, we show failure cases of our caption-and-box model.
    From the captions and object bounding boxes, the model cannot reliably infer the object states and appearances.
    Therefore, it fails to provide correct answers.
    However, visual embeddings, in principle, should capture these visual characteristics.
    We highlight the importance of devising MLLMs that can efficiently and effectively utilize distributed visual representations.

\begin{figure*}[t]
        \centering
        \includegraphics[width=\linewidth]{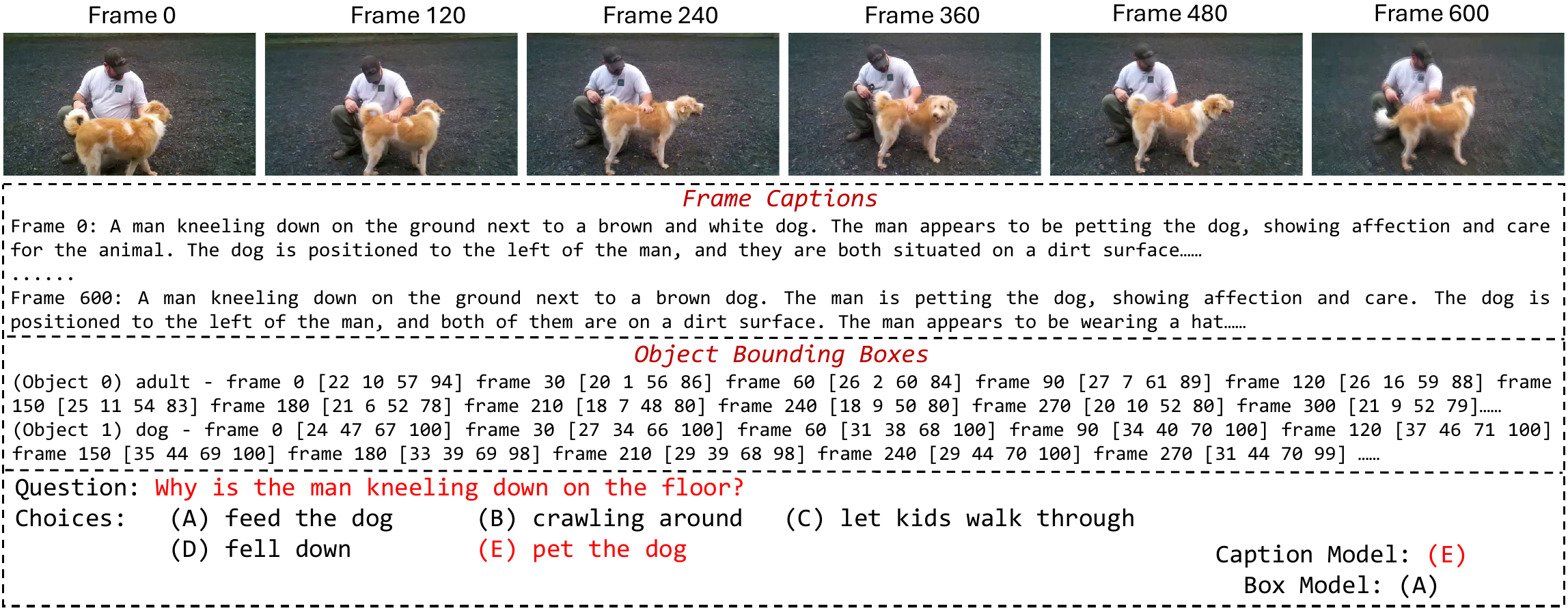}
        \vspace{-20pt}
        \caption{Failure case on NExT-QA. Although the detection and tracking algorithm can tell that there are an adult and a dog in the video, their actions cannot be inferred from the object bounding boxes. The captioning model can capture the person's action so that the model with captions as inputs correctly answers this question.}
        \label{fig:q_nextqa_fail1}
        \vspace{-5pt}
    \end{figure*}

    We also examine the failure cases on NExT-QA and IntentQA, where we found that our model struggles with questions involving human actions.
    For example, in \Cref{fig:q_nextqa_fail1}, the model with bounding box inputs is aware that there are a person and a dog in the video.
    However, the person's action cannot be determined from the bounding boxes.
    In contrast, as captions provide explicit descriptions of actions, the model with caption input is better at answering these questions, contributing to the performance gap in \Cref{tab:ablation_modality} compared to the box-only model.

\section{Unsuccessful Attempts}
\label{app:unsuccessful}

    \begin{table}[t]
    \centering
    \scalebox{0.80}{
        \begin{tabular}{cc}
            \toprule
            Box adapter &  Perception Test  \\
            \cmidrule(r){1-2}
            Language-based Representation  & 63.5 \\
            Embedding Projector & 60.1  \\
            Visual Prompting & 59.7  \\
            \bottomrule
        \end{tabular}
        }
    \vspace{-5pt}
    \caption{Performance of integrating bounding boxes using visual prompting. It is significantly less performant than the language-based representation and embedding projector.}
    \label{tab:visual_prompt}
    \end{table}
    
\subsection{Integrating boxes via visual prompting}
\label{app:visualized_bbox}
\begin{figure}[t]
        \centering
        \includegraphics[width=\linewidth]{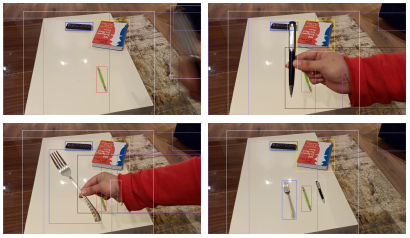}
        \vspace{-20pt}
        \caption{Examples of integrating bounding boxes via visual prompting from Perception Test.}
        \label{fig:visual_prompt}
        \vspace{-5pt}
    \end{figure}

Beyond integrating object-centric information via structural bounding box coordinates, we explore incorporating box information through visual prompting. Inspired by prior work~\cite{shtedritski2023does}, we overlay bounding boxes directly onto video frames and extract visual embeddings from these annotated frames. Within the same video, objects are distinguished by unique colors, and the color assigned to each object remains consistent across frames to maintain temporal coherence. Figure~\ref{fig:visual_prompt} demonstrates an example of the annotated frames from Perception Test. As shown in Table~\ref{tab:visual_prompt}, integrating bounding boxes using visual prompting performs worse than the embedding projector and language-based representation on Perception Test.

\subsection{Integrating object-level visual embeddings}
    
    \begin{table}[t]
    \centering
    \scalebox{0.8}{
        \begin{tabular}{cc|cc}
            \toprule
            Box & Visual embedding & Perception Test Acc. \\
            \midrule
            \checkmark & \ding{55} & 63.5 \\
            \checkmark & Frame-level & 62.7 \\
            \checkmark & Object-level & 63.3 \\
            \bottomrule
        \end{tabular}}
    \vspace{-5pt}
    \caption{Ablation on different visual embedding levels. Object-level visual embedding works better than the frame-level embedding but still cannot bring additional improvement when symbolic boxes are used.}
    \vspace{-10pt}
    \label{tab:object_clip_embedding}
    \end{table}

    Intuitively, fine-grained object appearances like texture cannot be accurately described by video frame captions and bounding boxes.
    But, in principle, they can be captured by distributed visual representations like CLIP \cite{Radford2021LearningTV} embeddings.
    However, \Cref{tab:ablation_modality} illustrates that integrating frame-level visual embeddings upon captions and bounding boxes does not bring additional benefits to the performance.

    Initially, we hypothesize that frame-level visual embeddings are too high-level to capture object details.
    To investigate this problem, we experiment with an object-level visual representation to replace the frame-level embedding.
    Specifically, we crop the objects from the video frames and extract their CLIP embeddings as object embeddings.
    Then, based on the language-based representation, we append each object embedding after its bounding box of the corresponding timestamp using the template below, where each $\langle|$obj\_emb$|\rangle$ indicates an object embedding.
    \begin{tcolorbox}[colback=green!5!white, colframe=green!50!black]
    (Object 0) bag – frame 0 [8 0 54 93] $\langle|$obj\_emb$|\rangle$ frame 90 [4 0 52 94] $\langle|$obj\_emb$|\rangle$ ......
    \end{tcolorbox}

    The results are in \Cref{tab:object_clip_embedding}.
    The object-level visual embedding outperforms the frame-level embedding but still falls short of the box-only model.
    This experiment again highlights the difficulty of integrating distributed embedding into MLLMs in a data-efficient manner, which would be a challenging but valuable research topic.

\end{document}